\documentclass{bmvc2k}

\title{Towards Efficient Neural Scene Graphs\\by Learning Consistency Fields}

\addauthor{Yeji Song}{ldynx@snu.ac.kr}{1}
\addauthor{Chaerin Kong}{veztylord@snu.ac.kr}{1}
\addauthor{Seoyoung Lee}{seoyoung1215@snu.ac.kr}{2}
\addauthor{Nojun Kwak$^*$}{nojunk@snu.ac.kr}{1}
\addauthor{Joonseok Lee$^*$}{joonseok@snu.ac.kr}{2,3}

\addinstitution{
 Department of Intelligence and Information,\\
 Seoul National University\\
 Seoul, Korea
}
\addinstitution{
 Graduate School of Data Science,\\
 Seoul National University\\
 Seoul, Korea
}
\addinstitution{
 Google Research\\
 Mountain View, CA, USA
}

\runninghead{SONG ET AL}{TOWARDS EFFICIENT NSG BY LEARNING CONSISTENCY FIELDS}

\usepackage{amsmath}
\usepackage{amssymb}
\usepackage{kotex}
\usepackage{multirow}
\usepackage{booktabs}
\usepackage{color}
\usepackage{capt-of}
\usepackage{wrapfig}
\usepackage{adjustbox}
\usepackage{hyperref}

\begin{document}

\maketitle
\def\thefootnote{*}\footnotetext{corresponding authors}

\begin{abstract}
Neural Radiance Fields (NeRF) achieves photo-realistic image rendering from novel views, and the Neural Scene Graphs (NSG) \cite{ost2021neural} extends it to dynamic scenes (video) with multiple objects. Nevertheless, computationally heavy ray marching for every image frame becomes a huge burden. In this paper, taking advantage of significant redundancy across adjacent frames in videos, we propose a feature-reusing framework. From the first try of naively reusing the NSG features, however, we learn that it is crucial to disentangle object-intrinsic properties consistent across frames from transient ones. Our proposed method,  \textit{Consistency-Field-based NSG (CF-NSG)}, reformulates neural radiance fields to additionally consider \textit{consistency fields}. With disentangled representations, CF-NSG takes full advantage of the feature-reusing scheme and performs an extended degree of scene manipulation in a more controllable manner. We empirically verify that CF-NSG greatly improves the inference efficiency by using 85\% less queries than NSG without notable degradation in rendering quality. Code will be available at \href{https://github.com/ldynx/CF-NSG}{https://github.com/ldynx/CF-NSG}.
\end{abstract}

\begin{figure}[h!]
  \centering
  \includegraphics[width=.95\linewidth]{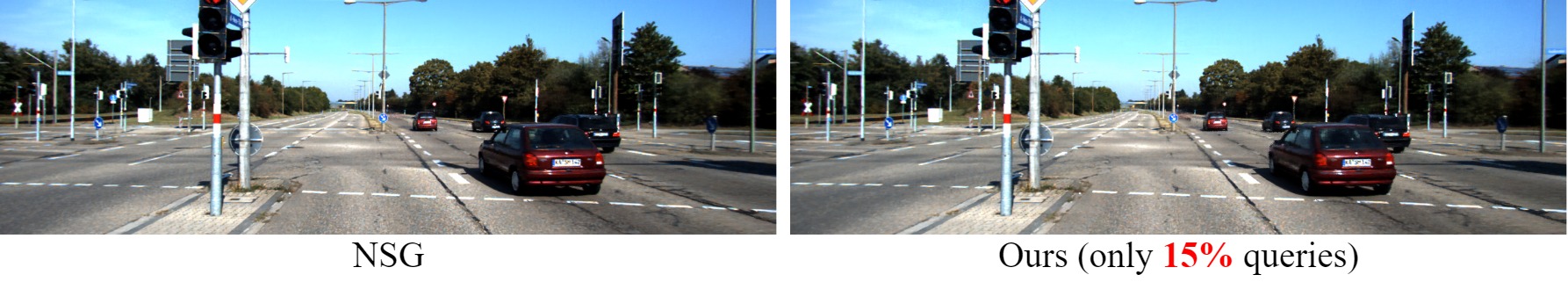}
  \caption{\textbf{Comparison of the qualitative results from NSG~\cite{ost2021neural} and our CF-NSG.} CF-NSG greatly improves the efficiency by using only 15\% of the original queries.}
  \label{fig:introduction2}
\end{figure}

\section{Introduction}

Being a long standing issue in computer vision, various approaches have been proposed for the task of novel view synthesis, including those using point clouds, discretized voxel grids and textured mesh \cite{fan2017point,yan2016perspective,sitzmann2019deepvoxels}. Neural Radiance Fields (NeRF) \cite{mildenhall2020nerf} represents a static scene by evaluating a volumetric density (transparency) and a view-dependent color. The object’s details can be expressed at a low storage cost via an implicit 5D (spatial location ($x, y, z$) and viewing direction ($\theta, \phi$)) function implemented with a multi-layer perceptron (MLP).

A similar idea is applied to understand a video scene. Unlike a still image, a video has an additional temporal-axis, and objects in the scene may be static or dynamic (moving) across time. Neural Scene Graphs (NSG) for dynamic scenes~\cite{ost2021neural} provides a considerable potential on various applications by enabling the understanding of a \textit{complex scene with dynamic multi-objects}, which has been tricky to model. Also, NSG extends the task of novel view synthesis to \textit{novel scene manipulation}, allowing spatial rearrangements of dynamic objects in the scene (\emph{e.g.}, the second row in  Fig.~\ref{fig:main_qualitative}). However, direct application of NSG is still not practical due to heavy computational overhead at inference. Along with other NeRF variants, NSG also suffers from expensive ray marching for all sampled queries for rendering. Moreover, since NSG repeats this procedure for each image frame independently, the total inference cost scales with the spatio-temporal resolution of the video. Observing this limitation, a natural question arises: how can we efficiently reduce this computational overhead?

We start from an idea that most objects in a video do not significantly change in adjacent frames. In other words, there is much redundancy in the representation of the scene across image frames as visual semantics are largely consistent. Reusing the previously-computed representations, instead of repeating computations for similar queries again, would be a plausible way to significantly improve efficiency. 

\begin{figure}[t]
  \centering
  \includegraphics[width=1\linewidth]{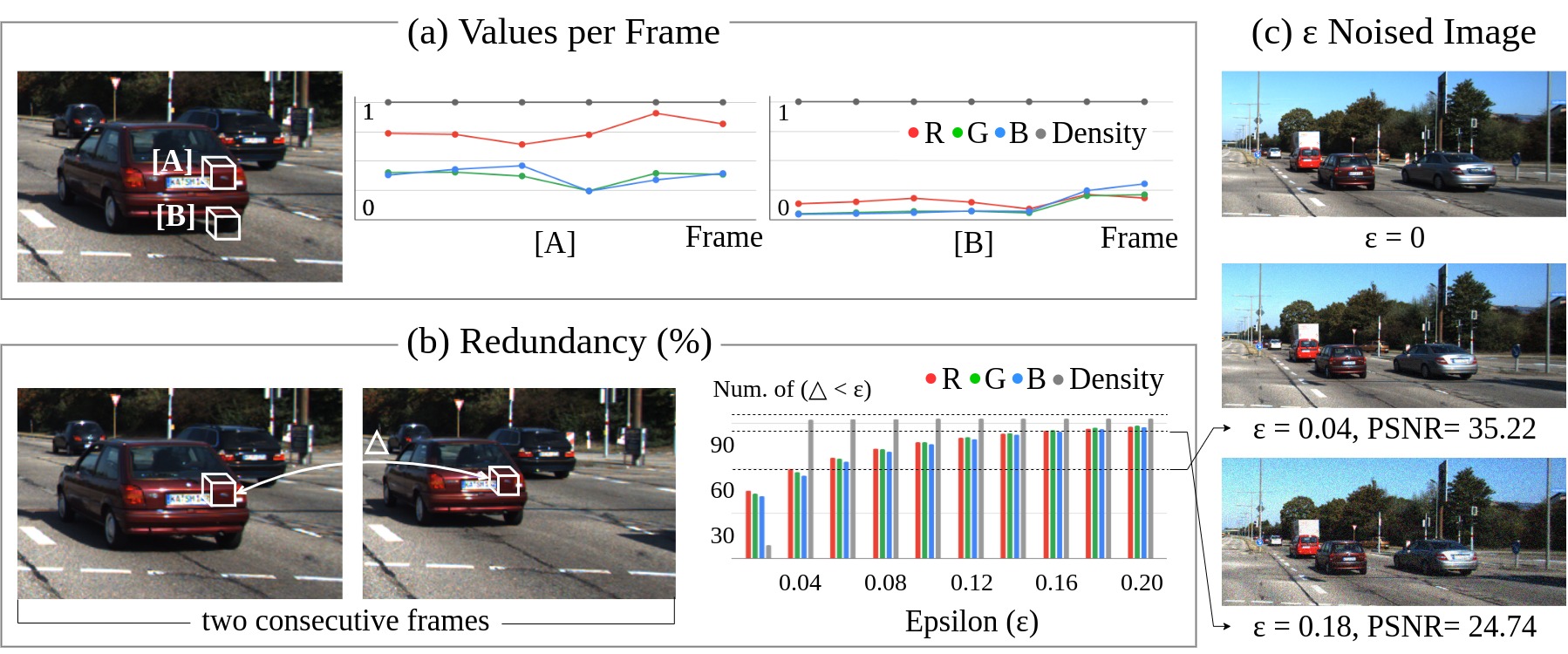}
  \caption{\textbf{Visualization of large redundancy in NSG~\cite{ost2021neural}.} \textbf{(a), (b):} Most of estimated color and density values are hardly or only slightly change across frames. \textbf{(c):} We add salt-and-pepper noise of $\epsilon$ to an image proving a majority of bins in (b) preserve similar values.}
  \label{fig:introduction1}
\end{figure}

We conduct a simple experiment to measure redundancy across frames. We divide an object's bounding box into uniform spatial bins and assign queries $(x, y, z)$ to each bin, storing the estimated color and density values.
Fig.~\ref{fig:introduction1}(a) shows the values across all frames for two selected bins {\small \textbf{[A]}} and {\small \textbf{[B]}}. We found that the color and density values from the consecutive frames hardly or only slightly change within a limited range. Fig.~\ref{fig:introduction1}(b) shows the ratio of all bins whose values change less than a threshold $\epsilon$ in two consecutive frames. As the range of radiance and density are within $[0, 1]$, the possible threshold for $\epsilon$ is also within $[0, 1]$. We note that redundancy is already close to 70\% at $\epsilon \approx 0.04$. To facilitate better understanding of $\epsilon $, we generate images with random salt-and-pepper noise of $\epsilon$ = $\{0.04, 0.18\}$ for all pixels. As shown in Fig.~\ref{fig:introduction1}(c), the image with $\epsilon=0.04$ noise is almost identical to the original image maintaining high PSNR. The image with $\epsilon=0.18$ noise (90\% redundancy in (b)) still well-preserves the overall scene. This experiment indicates that most of bins share highly similar values across frames.
Hence, actively leveraging this redundancy is not only beneficial but in fact necessary to guide our model to learn dynamic scenes more efficiently.

To this end, we propose to identify visual components that are consistent across frames, and reuse them for improved efficiency. Fig.~\ref{fig:introduction2} shows the rendered image using our method with 85\% less queries without significantly degrading rendering quality. In this paper, we propose Consistency-Field-based NSG (CF-NSG), which effectively reduces redundant computation across frames for more efficient rendering. We first review NSG in Sec.~\ref{sec:Preliminaries}, and analyze a critical factor that hinders the feature reusing mechanism in Sec.~\ref{sec:Naive}. Then, we introduce our CF-NSG in Sec.~\ref{sec:Proposed}. We empirically demonstrate that our model greatly improves efficiency with no distinguishable compromise in image quality in Sec.~\ref{sec:Experiments}.

\section{Preliminaries} \label{sec:Preliminaries}

\textbf{Neural Radiance Fields (NeRF)} \cite{mildenhall2020nerf} captures 3D representations of static objects or scenes. The implicit representation is encoded into an MLP, which maps a 3D spatial location $\textbf{\textit{x}} = (x, y, z)$ to its volumetric density $\sigma$, combined with a viewing direction $\textbf{\textit{d}} = (\theta, \phi)$ to an emitted color $\textbf{\textit{c}} = (r,g,b)$. The color of the pixel $C(\textbf{\textit{r}})$ along a camera ray $\textbf{\textit{r}}$ is estimated by accumulating the color and transmittance of $N$ sampled points along the ray:

\begin{equation} \label{eq:nerf_vol_rend}
  \footnotesize
  \hat{C}(\textbf{\textit{r}}) = \sum_{i=0}^{N-1} T_i \left( 1-\exp(-\sigma_i \delta_i) \right) \textbf{\textit{c}}_i, \quad
  T_i = \exp \Bigg(- \sum_{j=0}^{i-1} \sigma_j \delta_j \Bigg)
\end{equation}
where $\delta_i$ is the distance between adjacent sampled points. The NeRF network is optimized to reduce the $L_2$ distance between the estimated colors $\hat{C}(\textbf{\textit{r}})$ for a random 
a batch of rays $\mathcal{R}$ and their ground truth (GT); that is, $  \mathcal{L} = \sum_{\textbf{\textit{r}} \in \mathcal{R}} 
     \| C(\textbf{\textit{r}}) - \hat{C}(\textbf{\textit{r}}) \|_2^2 
  $.
 
\noindent
\textbf{Neural Scene Graphs (NSG)}~\cite{ost2021neural} classifies objects in a multi-object dynamic scene (video) according to whether they are moving or static across frames. They define the mappings from global coordinates to each dynamic object's canonical coordinates using transformations of the object (\textit{e.g.,} translation, rotation and scaling). They also group dynamic objects into classes (\textit{e.g.,} car, pedestrian, cyclist), defining a common canonical coordinates of each class. Meanwhile, all the static objects are grouped as the background. Then, canonical coordinates of the each class as well as the background are represented using separate NeRFs. 

The overall process can be expressed by
\begin{equation} \label{eq:NSG}
    \small
    \mathcal{F}_{\theta_{bg}}: (\textbf{\textit{x}}, \textbf{\textit{d}}) \rightarrow (\textbf{\textit{r}}, \textbf{\textit{g}}, \textbf{\textit{b}}, \sigma), \quad \quad
    \mathcal{F}_{\theta_{c}}: (\textbf{\textit{x}}_o, \textbf{\textit{d}}_o, \textbf{\textit{p}}_o, \textbf{\textit{l}}_o) \rightarrow (\textbf{\textit{r}}, \textbf{\textit{g}}, \textbf{\textit{b}}, \sigma),
\end{equation}
where $\theta_{bg}$, $\theta_{c}$ are respective weights for static and dynamic representation models.
Here,  $c$ denotes each class of dynamic objects, where $c \in \{c_1, \cdots, c_n\}$ for $n$ classes.
In the dynamic model, the set of dynamic objects belonging to class $c$ ($\mathcal{O}_c$) shares weights $\theta_{c}$, while latent vector $\textbf{\textit{l}}_{o}$ is uniquely learned for an individual object $o \in \mathcal{O}_c$.  $\textbf{\textit{p}}_{o}$ is the global location of object $o$ in the scene. Each $\textbf{\textit{x}}_{o}$ and $\textbf{\textit{d}}_{o}$ refer to the spatial location and the viewing direction, respectively, where the corresponding color values are desired. Note that $\textbf{\textit{x}}_{o} \in [-1, 1]^3$ and $\textbf{\textit{d}}_{o}$ are in the canonical coordinates of object $o$, different from $\textbf{\textit{x}}$ and $\textbf{\textit{d}}$ in global coordinates. Outputs $(r, g, b, \sigma)$ from the dynamic models are mapped from canonical coordinates to global coordinates and integrated simultaneously with the outputs from the static model along the rays, composing a scene similarly as in~\cite{niemeyer2021giraffe}.

\section{First Try: A Naive Reuse of NSG Features}
\label{sec:Naive}

\begin{figure}[t]
  \centering
  \includegraphics[width=1\linewidth]{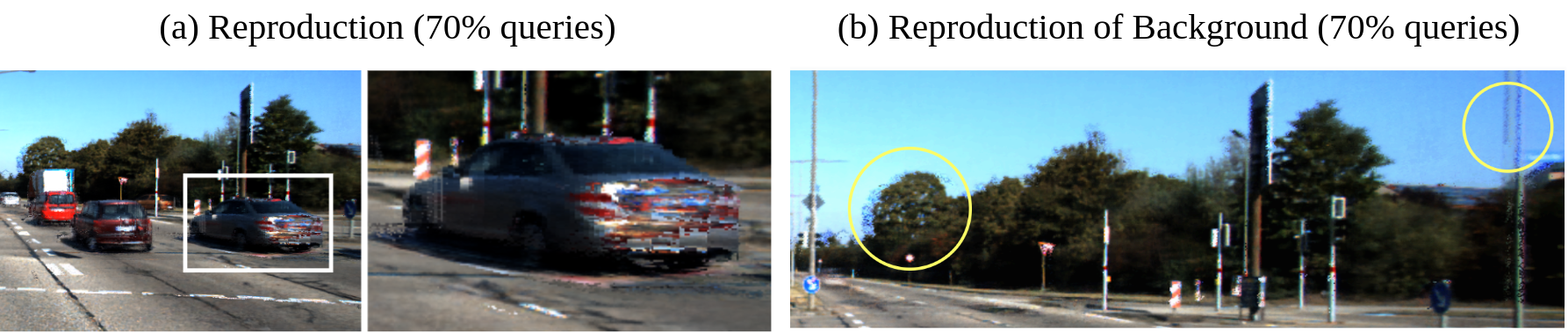}\\
  \caption{\textbf{Rendering results with naive reusing of NSG.} We reproduce a frame seen during training with a naive reuse of NSG features, resulting in degradation of image quality.}
  \label{fig:naive}
\end{figure}

NSG extends the realm of NeRF-based applications to videos with multiple dynamic objects. However, computation cost at inference is still prohibitive, as image rendering requires ray marching operations per frame. Thus, we propose a \textit{feature-reusing framework} that stores redundant features across frames during training, then at inference reuses them instead of going through a full forward pass. Assuming the features in the nearby voxels are similar, we create quantized memory bins in the coordinates as follows. For the dynamic models ($\mathcal{F}_{\theta_{c}}$ in Eq.~\eqref{eq:NSG}), we quantize the object-specific 3D coordinates (within the object's bounding box) into spatial bins, similarly as in the experiment in Fig.~\ref{fig:introduction1}. For the static model ($\mathcal{F}_{\theta_{bg}}$ in Eq.~\eqref{eq:NSG}), following NSG, we define radiance fields and grid memory bins on a set of 2D planes instead of volume. If a query is assigned to the bin whose value is expected not to significantly change across frames, we reuse previously computed results from the bin.

\noindent
\textbf{A Naive Approach of Reusing NSG features.}
At the beginning, we use a pretrained NSG model to compute the color (RGB) and density ($\sigma$) for the training set, caching them in the memory bins. To estimate the expected change of each bin across frames, we additionally store the gradients of color and density with respect to global location $\textbf{\textit{p}}_{o}$, viewing direction $\textbf{\textit{d}}_o$ and $\textbf{\textit{d}}$, inputs that may change across frames. A bin with a small gradient indicates that values in that bin are most likely not to change across frames.Therefore, we examine whether queries are assigned to the bins whose gradient norm is smaller than a predetermined threshold ${\tau}_{\partial}$ at inference. For such queries, we directly retrieve the stored values from the bins. Otherwise, the full MLP path is used to compute new RGB and $\sigma$. We apply gradient norm {\small $\partial = \|\partial{\omega} / \partial{\textbf{\textit{p}}_o}\|^2_2 + \|\partial{\omega} / \partial{\textbf{\textit{d}}_o}\|^2_2$} for the dynamic models, and {\small $\partial = \|\partial{\omega} / \partial{\textbf{\textit{d}}}\|^2_2$} for the static model, where {\small $\omega \in \{r, g, b, \sigma\}$}. Fig.~\ref{fig:method}(b) presents an overall naive reusing process.

Fig.~\ref{fig:naive} illustrates the qualitative results. Fig.~\ref{fig:naive}(a) shows an example of a reproduced dynamic object, a vehicle. It clearly shows that reusing RGB color values generates abnormalities, \textit{e.g.}, blended colors at the rear of the grey car. The naive reusing approach is also imperfect for reproducing a static background. As seen in Fig.~\ref{fig:naive}(b), the street lamp on the right or trees in the back (yellow circled) cannot be pinned down, displaying ghost effects. This indicates the reusing decision was made erroneously, reusing the features that are significantly changes across the frames.

\noindent
\textbf{Limitation of the NSG Representations.}
From the previous example, we notice two inadequacies in naive reusing of NSG that may lead to the disappointing results. First, gradient norm may not provide a convincing criterion for which values in the bins are reusable and advantageous for the rendering quality. Second, RGB color values may not be appropriate for direct reuse. As shown in Fig.~\ref{fig:introduction1}, RGB and $\sigma$ hardly change in most frames. Then, what factors do cause these inadequacies? We notice that RGB values change abruptly once in a while due to external factors such as shadow from nearby environment, global illumination, or the change of a dynamic object's location. Considering features that are independent of these external factors would lead to a more reliable criterion as well as an appropriate reuse.

First, we define two distinct properties that are canonical to the object (\emph{e.g.}, intrinsic color~\cite{zhang2019vehicle} or shape) and those from external factors. The former should be maintained consistent across frames regardless of the object's current location or viewing direction, while the latter may change depending on the spatio-temporal environment. Here, spatio-temporal consistency incorporates both actual time change or movement of an object (temporal change) and the viewpoint change coming from the use of stereo-cameras on different locations (spatial change). Therefore, we aim to reuse features representing canonical properties.

\begin{figure}[t]
  \centering
  \includegraphics[width=1\linewidth]{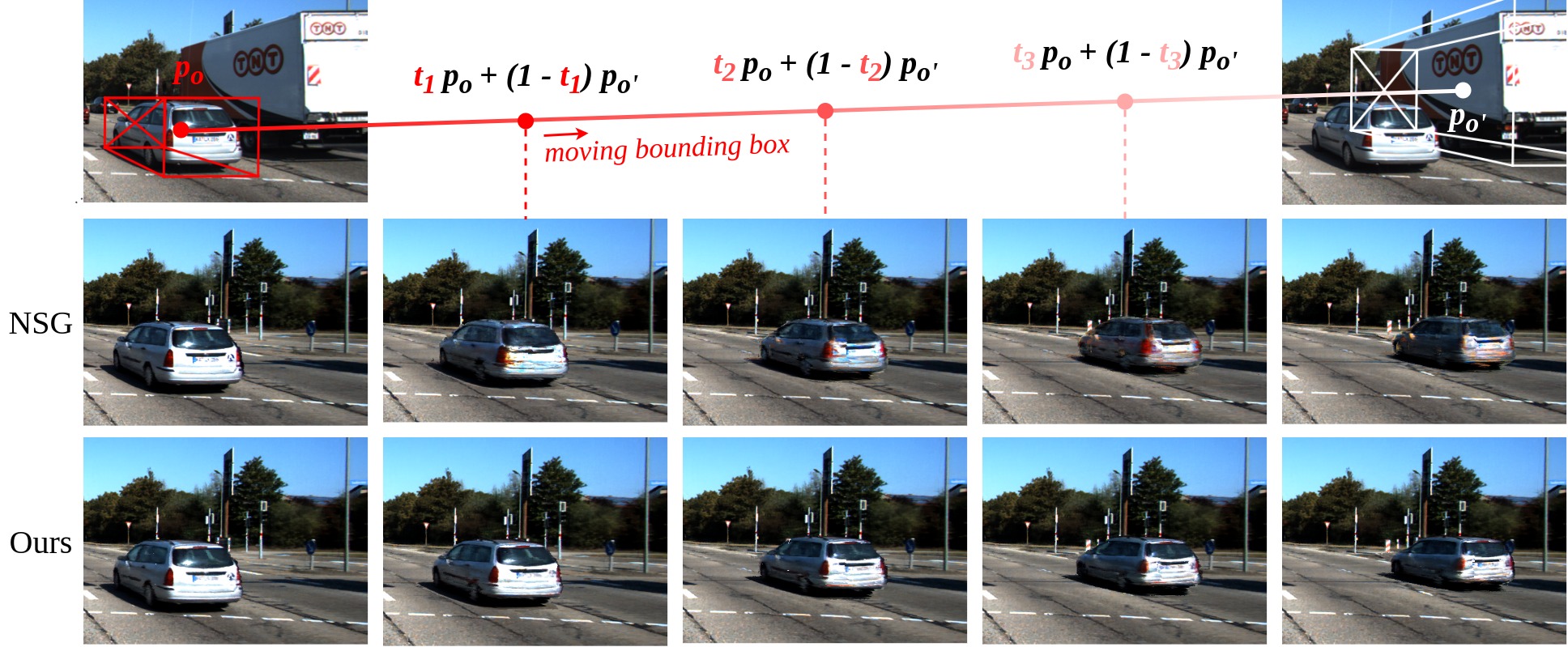}
  \caption{\textbf{Translation experiment of a dynamic object.} In NSG, an object's color and shape of tail light change depending on the object's global location. Meanwhile, CF-NSG well preserves canonical properties of the object disentangled from environmental factors.}
 \label{fig:entanglement}
\end{figure}

However, we hypothesize that NSG fails to completely disentangle features intrinsic to each object and those affected by its transient environment. Since RGB color values are produced by both object-internal and external factors, supervision from RGB color (GT pixel values) alone would be unable to instruct about canonical properties of the object. To support our hypothesis, we conduct an experiment shown in Fig.~\ref{fig:entanglement}. The first row shows two objects, a white car ($o$) and a truck behind it ($o'$) in the training set, whose bounding boxes $\textbf{\textit{p}}_{o}$ and $\textbf{\textit{p}}_{o'}$ are marked in red and white, respectively. In the second row, we render the white car ($o$) at intermediate locations between $\textbf{\textit{p}}_{o}$ and $\textbf{\textit{p}}_{o'}$, using the NSG representation of the car ($\textbf{\textit{l}}_{o}$). Even though we only change its physical location, we observe that canonical properties of the object, \emph{e.g.}, its color and shape of the tail lights, are significantly contaminated depending on its location. This result reveals that the object-intrinsic and environmental representations are severely entangled in NSG. That is, the object-intrinsic representation learned by NSG is heavily affected by transient factors, such as the viewing direction or its location.

\section{The Proposed Method: Consistency-Field-based NSG} \label{sec:Proposed}

To improve the efficiency of NSG by the feature-reusing framework while maintaining the image quality, our observations from Sec.~\ref{sec:Naive} imply two conditions: 1) a solid criterion should be used to determine which bins are reusable without hurting the rendering quality, and 2) the reused features should be able to properly represent the characteristics that is intrinsic to each object. We propose the Consistency-Field-based NSG (CF-NSG), illustrated in Sec.~\ref{subsec:CF-NSG} and Fig.~\ref{fig:method}, satisfying these two conditions and achieving disentangled representations. By the term \textit{consistency}, we refer to the characteristics of the query that strongly show canonical and consistent properties of the object across frames. We call our method \textit{Consistency-Field} since CF-NSG reformulates the radiance fields to additionally measure the consistency of each query across frames. We also discuss how we further boost the efficiency and adapt our final model to practical settings under limited memory in Sec.~\ref{subsec:Further}.

\subsection{Consistency Fields and Reusable Features} \label{subsec:CF-NSG}

\begin{figure}[t]
  \centering
  \includegraphics[width=1\linewidth]{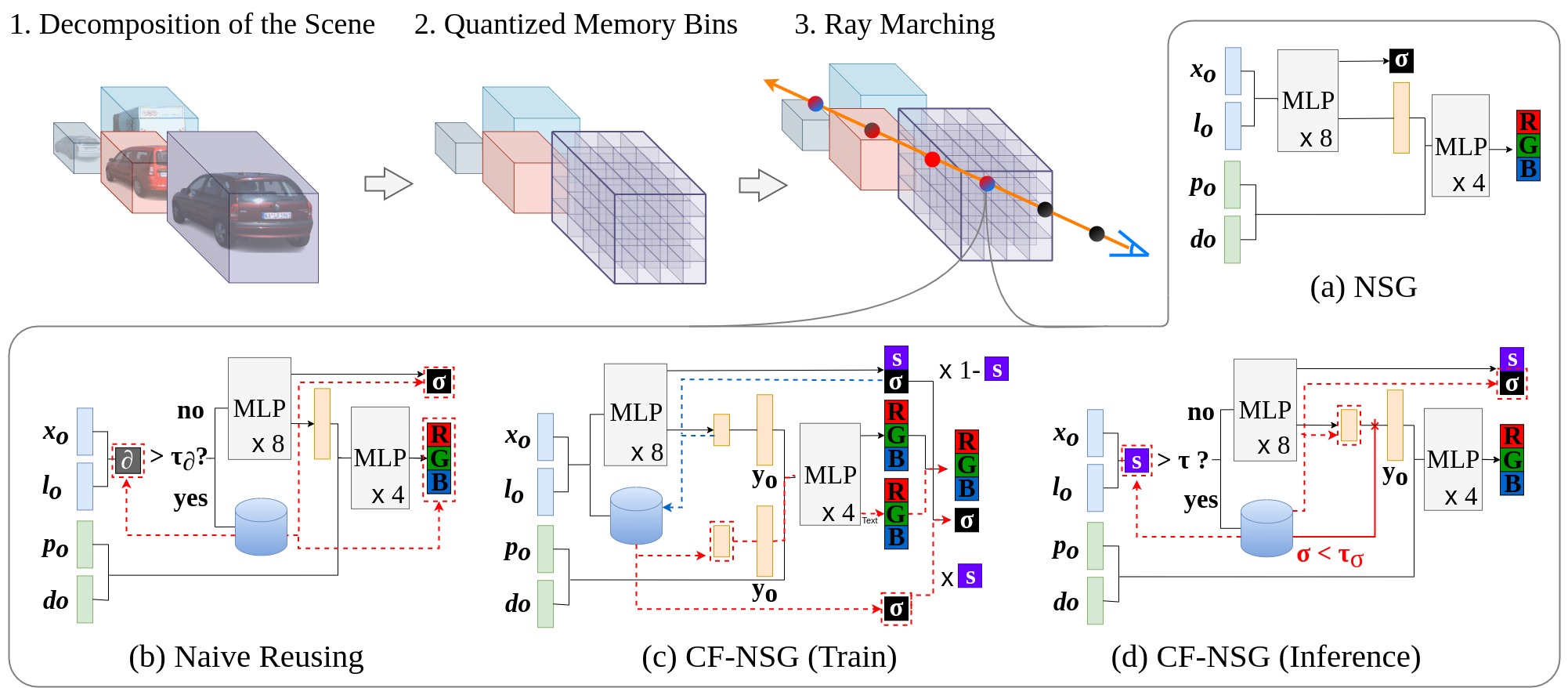}\\
  \caption{\textbf{The overall pipeline of CF-NSG (the dynamic model).} \textbf{(a)--(d):} Instead of computing the whole queries, CF-NSG utilizes feature-reusing frameworks reducing an amount of full feed-forward pass at inference. \textbf{(c):} CF-NSG learns consistency fields and appropriately reuses features by loss from interpolated outputs (see Eq.~\ref{eq:aggregation}) and regularizer (see Eq.~\ref{eq:Ours})}
  \label{fig:method}
\end{figure}

\noindent
\textbf{Solid Reusability Criterion.}
We have shown in Sec. \ref{sec:Naive} that the gradient-norm-based naive reusing cannot retain the quality of rendered images. Therefore, we take a learnable approach to enable our model to establish a stronger criterion. Compared to NSG in Fig.~\ref{fig:method}(a), our CF-NSG in Fig.~\ref{fig:method}(c-d) additionally returns \textit{consistency scores} $s \in [0, 1]$ estimating how consistent the query is across frames. We indicate $s_{bg}(\textbf{\textit{x}})$ for the static model, and $s_{c}(\textbf{\textit{x}}_o)$ for the dynamic model of object $o$ belonging to class $c$, respectively. The queries with higher consistency scores than a predetermined threshold $\tau$ have a strong tendency to be invariant spatio-temporally, meaning that they are safe to be reused in other frames.

Since there is no ground truth for consistency scores, we give an auxiliary loss at training as follows: both the full feed-forward pass (\textbf{\small \textcolor{black}{black solid line}} in Fig.~\ref{fig:method}(c)) and the reuse pass (\textbf{\small \textcolor{red}{red dashed line}}) are activated, producing $\omega_\text{full}$, $\omega_\text{reuse} \in \{(r, g, b, \sigma) : r, g, b, \sigma \in [0, 1]\}$, respectively. Then, they are aggregated to
\begin{equation}
  \small
  \omega_\text{mixed} = s \cdot \omega_\text{reuse} + (1 - s) \cdot \omega_\text{full}
  \label{eq:aggregation}
\end{equation}
A batch of $\omega_\text{mixed}$ are integrated along rays to yield an interpolated image. The color difference between this interpolated and GT images induces the loss, which is backpropagated to update $s$ as well as other parameters. At inference, we choose each query's path based on $s$, to reuse (\textbf{\small \textcolor{red}{red dashed line}} in Fig.~\ref{fig:method}(c)), skip (\textbf{\small \textcolor{red}{red solid line}}, see Sec.~\ref{subsec:Further}) or full feed-forward. 

\noindent
\textbf{Reusable Representations.}
As observed in Sec.~\ref{sec:Naive}, the quality of the rendered images deteriorates as inappropriate features are reused. To overcome this issue, we explicitly learn a \textit{canonical feature} $\mathbf{y} \in \mathbb{R}^{l}$, where $l$ is a hyperparameter.
$\mathbf{y}_{bg}$ indicates that of the static model and is a function of $\textbf{\textit{x}}$. The canonical feature $\mathbf{y}_o$ of a dynamic object $o$ is a function of its latent vector $\textbf{\textit{l}}_o$ along with $\textbf{\textit{x}}_{o}$. These canonical features differ from the intermediate features of NSG. NSG uses intermediate features to map the positional and directional inputs to higher-frequency, whereas our canonical features are for well-representing spatio-temporarily consistent properties through the feature-reusing framework.
It is these canonical features $\textbf{y}$ (along with the density $\sigma$) that are stored in the corresponding memory bin at training (\textbf{\small \textcolor{blue}{blue dashed line}} in Fig.~\ref{fig:method}(c)), contrast to the actual RGB values used in the naive reusing in Sec.~\ref{sec:Naive}, Fig.~\ref{fig:method}(b). Also, reusing is based on consistency score instead of gradient norm $\partial$ at inference.

\noindent
\textbf{Overall Pipeline.}
Fig.~\ref{fig:method}(c-d) illustrate the overall pipeline of our CF-NSG for the dynamic model. CF-NSG learns consistency score $s$ and canonical features $\mathbf{y}_o$ simultaneously, and reuses $\mathbf{y}_o$ depending on $s$. If the model reuses improper features largely affected by transient factors, $L_2$-loss between the rendered image and GT would increase and imposes a penalty. Thus, $\mathbf{y}_o$ with a high $s$ are encouraged to be even more object-intrinsic across frames. The static model also follows a similar structure, illustrated more in detail in the Supp. Materials. We also provide pseudocodes for both dynamic and static models in the Supp. Materials.

\noindent
\textbf{Training Objective.}
Our full objective is as follows:
\begin{equation} \label{eq:Ours}
\footnotesize
    \mathcal{L} = \sum_{\textbf{\textit{r}} \in \mathcal{R}} \Big[ \big\|\hat{C}(\textbf{\textit{r}}) - C(\textbf{\textit{r}}) \big\|_2^2 + \big\| \hat{C}_\text{mixed}(\textbf{\textit{r}}) - C(\textbf{\textit{r}}) \big\|_2^2\Big] + \sum_{\textbf{\textit{x}}, \textbf{\textit{x}}_o \in \mathcal{X}} \Big[  \frac{1}{\|s_{bg}(\textbf{\textit{x}})\|^2} + \frac{1}{\|s_c(\textbf{\textit{x}}_o)\|^2} \Big] + \frac{1}{v} \|\textbf{\textit{l}}_o\|_2^2,
\end{equation}
where we denote the predicted pixel color from a ray \textbf{\textit{r}} by full feed-forward rendering as $\hat{C}$, the one by aggregation in Eq.~\eqref{eq:aggregation} as $\hat{C}_\text{mixed}$, and the reference color as $C$. 
The original training objective of NSG consists of the $L_2$-loss on the rendered image (the first term in Eq.~\eqref{eq:Ours}) and a Gaussian prior on the object latent vector~\cite{park2019deepsdf} (the last term). To encourage our model to reason about consistency, we introduce additional regularization to the objective function. Since we aggregate $\omega_\text{mixed}$ by a convex combination of full feed-forward outputs and reused outputs, the model might converge to the trivial solution of zero reuse. Thus, we add a regularization term to penalize low reuse scores for $s_{bg}$ and $s_{c}(\textbf{\textit{x}}_o)$ (the second term).

\noindent
\textbf{Where Does Disentanglement Come from?} 
In NSG, where $L_2$-loss between the rendered and GT images is the only supervision, intermediate features do not need to preserve unique information about an object or a scene from $\textbf{\textit{l}}_o$, $\textbf{\textit{x}}_o$ or $\textbf{\textit{x}}$, diluted with other inputs quickly. CF-NSG, on the other hand, is simultaneously trained by an additional task, estimating the consistency score $s$ to predict the proper reusability. Since the model produces both the canonical features $\textbf{y}$ and $s$, $\textbf{y}$ is naturally aware of consistent properties of the corresponding bin. For this reason, while NSG appears to confuse intrinsic vehicle properties and its transient location in Fig.~\ref{fig:entanglement}, CF-NSG is less affected thanks to better disentanglement between them. 

\noindent
\textbf{Are Environmental Effects Not Considered?} 
We emphasize that CF-NSG \underline{\textit{does not}} disregard transient factors and is able to learn environmental properties as well. The second MLP, a shallow 4-layer MLP, receives transient inputs (global location $\textbf{\textit{p}}_o$, viewing direction $\textbf{\textit{d}}_o$ and $\textbf{\textit{d}}$) along with $\mathbf{y}$ that well-represents object-intrinsic properties, enabling the second MLP to learn transient properties more effectively. This process is not skipped regardless of $s$.

\subsection{Further Improvements} \label{subsec:Further}

\textbf{Skipping.}
We utilize the consistency score $s$ more aggressively to further boost the efficiency. We find that a bin with a high consistency score $s$ and a low density $\sigma$ is likely to be an empty space without a relevant scene content. We skip such a bin and allow queries to be more densely distributed near the content. NSVF~\cite{liu2020neural} is closely related to our work in that they improve the efficiency of NeRF by defining a set of voxels in the scene and skipping the empty voxels. However, the difference arises from the way of finding the empty space. NSVF prunes the empty voxels based on $\sigma$, while our method uses both the $s$ and $\sigma$. Relying only on $\sigma$ requires a finer adjustment of the threshold between an almost transparent car window and a truly empty space. On the other hand, they can be distinguished through $s$; the former has a lower $s$, while the latter has a higher $s$, since the latter has consistently low density over time. Our method enjoys additional gain in efficiency by skipping more aggressively. 

\noindent
\textbf{Memory-efficient Implementation.}
Storing canonical features $\mathbf{y} \in \mathbb{R}^{l}$ for each spatial bin imposes a considerable memory footprint, making the framework impractical. That is, we are faced with two contradicting desiderata: keeping as little information in memory as possible, while mapping it to a rich feature space. To this end, we apply a learnable factorization inspired by \cite{skorokhodov2021adversarial,suarez2017language}.
Specifically, $\mathbf{y}$ is factorized to $\text{\small \textbf{flatten}}(\mathbf{u}_1 \times \mathbf{u}_2) + \mathbf{z}$, where $\mathbf{z} \in \mathbb{R}^l$, $\mathbf{u}_i \in \mathbb{R}^{m^2}$, and $\sqrt{l} = m^2$. Each $\mathbf{u}_i$ is similarly factorized again by
$\mathbf{u}_i = \text{\small \textbf{flatten}}(\mathbf{v}_{i,1} \times \mathbf{v}_{i,2})$, 
where $\mathbf{v}_{i,j} \in \mathbb{R}^m$.
Here, $\mathbf{z}$ keeps shared information about each component of the scene (\textit{e.g.,} each dynamic object or background) and is shared for all queries belonging to the same component. Since it is stored only once per each component, we keep $\mathbf{z}$ unfactorized to fully enjoy $l$ degree of freedom. On the other hand, the rest, memory-bin-specific components, are aggressively factorized to rank-$m$ for efficient memory footprinting. We use $l = 256, m = 4$. Instead of storing 256D, we store 4D $\times 4$, thereby reducing the memory usage by 93\%.

\section{Experiment} \label{sec:Experiments}

\noindent
\textbf{Implementation Details.}
We use KITTI~\cite{geiger2012we} and Objectron~\cite{ahmadyan2021objectron} datasets that provide 3D bounding boxes for objects in a scene. KITTI provides multi-object tracking information captured by stereo-cameras while Objectron contains more diverse types of objects with more drastic camera view changes. We set baselines including NSG~\cite{ost2021neural}, NeRF~\cite{mildenhall2020nerf} and NeRF with temporal inputs~\cite{ost2021neural}. We also compare with NSVF~\cite{liu2020neural} that improved efficiency of NeRF and D-NeRF~\cite{pumarola2021d} that extended NeRF to the dynamic scenes. However, since the former is mainly targeted to the static scenes and the latter is not considering efficiency, they are not directly comparable. Nevertheless, we include them as baselines to show a general tendency. We do not compare with NSFF~\cite{li2021neural} since the unbounded scene of KITTI induced an unstable training for depth estimation loss. We refer to Supp. Materials for more details.

\begin{figure}
  \centering
  \includegraphics[width=1\linewidth]{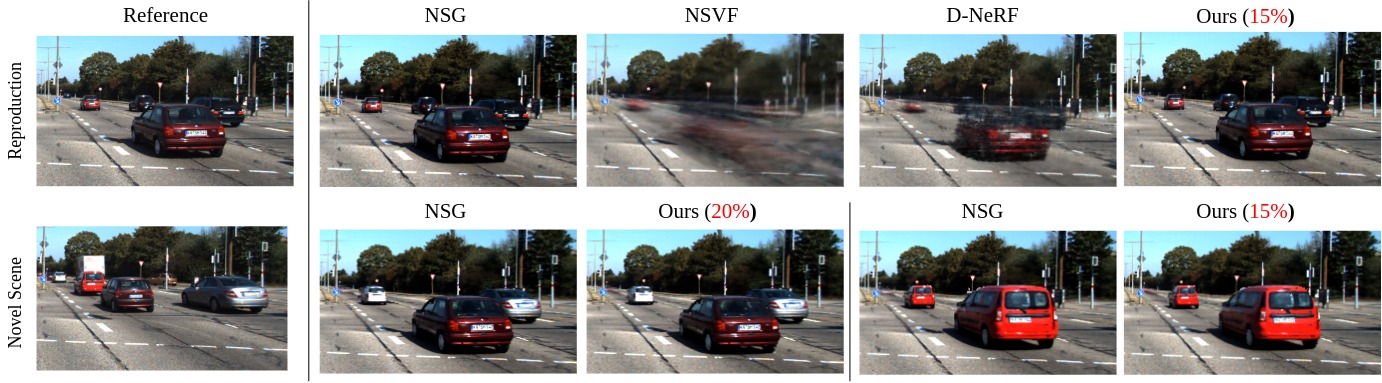}\\
  \caption{\textbf{Qualitative comparisons on KITTI~\cite{geiger2012we}}. We compare NSG~\cite{ost2021neural}, NSVF~\cite{sitzmann2019deepvoxels}, D-NeRF~\cite{pumarola2021d} and CF-NSG. We note that CF-NSG uses 80--85\% fewer number of queries.}
  \label{fig:main_qualitative}
\end{figure}

\noindent
\textbf{Comparison with Baselines.} 
Tab.~\ref{tab:main_table} quantitatively compares the quality of images from each implicit neural representation framework, with its computational cost indicated by the number of queries. Impressively, our CF-NSG achieves image quality close to that of NSG by only using 15--53\% of queries. Note that the table is sorted by the number of queries used by each method, so for the methods below NSG, NSG scores are considered as the \underline{\textit{upper bound}}. When we train vanilla NSG with reduced number of queries (`NSG-reduced' in Tab.~\ref{tab:main_table}), we observe significant performance drop.
Fig.~\ref{fig:main_qualitative} compares our method to baselines qualitatively. The first row compares our CF-NSG against baselines on reproduction of a frame seen during training, and the second row compares NSG and CF-NSG on novel scene manipulation, where we sample dynamic objects in the reference frame and generate a scene in a new arrangement. We can see that the rendered images of NSG and ours are almost indistinguishable from human eyes in spite of significantly less number of (15--20\%) quires.

\begin{table}[t]
    \centering
	\resizebox{1\textwidth}{!}{
	\begin{tabular}{cl|c||c|c|c|c|c}
	\toprule
	{\parbox{0.6cm}{\centering \small Dataset}} & 
	{\parbox{1.5cm}{\centering \small Method}} & {\parbox{1.5cm}{\centering \small \#Queries}} & {\parbox{1.6cm}{\centering \small PSNR($\uparrow$)}} & {\parbox{1.6cm}{\centering \small SSIM~\cite{wang2003multiscale}($\uparrow$)}} & {\parbox{1.6cm}{\centering \small LPIPS~\cite{zhang2018unreasonable}($\downarrow$)}} & {\parbox{1.6cm}{\centering \scriptsize tOF~\cite{chu2020learning}$\times10^6$($\downarrow$)}} & {\parbox{1.6cm}{\centering  \scriptsize tLP~\cite{chu2020learning}$\times100$($\downarrow$)}} \\
    \midrule
	\multirow{7}{*}{KITTI} & D-NeRF~\cite{pumarola2021d} & 8.72$\times$ & 16.33 & 0.505 & 0.418 & 3.823 & 4.835 \\
	& NeRF~\cite{mildenhall2020nerf} & 7.90$\times$ & 20.99 & 0.621 & 0.446 & 2.702 & 3.840 \\
	& NSVF~\cite{liu2020neural}
	& 6.38$\times$ & 22.95 & 0.706 & 0.386 & 2.831 & 5.071 \\
	& NeRF+time~\cite{ost2021neural} & 1.64$\times$ & 24.86 & 0.653 & 0.492 & 2.272 & 1.563 \\
	\cmidrule(lr){2-8}
	& NSG~\cite{ost2021neural} & 1$\times$ & 29.54 & 0.914 & 0.171 & 0.619 & 0.265 \\
	\cmidrule(lr){2-8}
	& NSG-reduced~\cite{ost2021neural} & 0.75$\times$ & 24.69 & 0.702 & 0.452 & 1.625 & 1.990 \\
	& CF-NSG (ours) & \textbf{0.15$\times$} & \textbf{28.70} & \textbf{0.891} & \textbf{0.204} & \textbf{0.766} & \textbf{0.266} \\
	\midrule
	\midrule
	\multirow{3}{*}{\parbox{1.3cm}{\centering Objectron \\ \textit{chair}}}
	& NSG~\cite{ost2021neural} & 1$\times$ & 29.20 & 0.864 & 0.286 &  0.224 & 0.411\\
	\cmidrule(lr){2-8}
	& NSG-reduced~\cite{ost2021neural} & 0.29$\times$ & 27.65 & 0.866 & 0.263 & 0.245 & \textbf{0.296} \\
	& CF-NSG (ours) & \textbf{0.28$\times$} & \textbf{28.58} & \textbf{0.878} & \textbf{0.259} & \textbf{0.229} & 0.354 \\
	\midrule
	\multirow{3}{*}{\parbox{1.3cm}{\centering Objectron \\ \textit{camera}}} 
	& NSG~\cite{ost2021neural} & 1$\times$ & 29.17 & 0.854 & 0.273 & 0.414 & 0.673 \\
	\cmidrule(lr){2-8}
	& NSG-reduced~\cite{ost2021neural} & 0.85$\times$ & 22.82 & 0.701 & 0.454 & 1.182 & 1.383 \\
	& CF-NSG (ours) & \textbf{0.53$\times$} & \textbf{26.01} & \textbf{0.789} & \textbf{0.363} & \textbf{0.471} & \textbf{0.634} \\
	\bottomrule
	\end{tabular}}
	\vspace{.1mm}
    \caption{\textbf{Quantitative Comparison on KITTI~\cite{geiger2012we} and Objectron~\cite{ahmadyan2021objectron}}. CF-NSG achieves comparable results with NSG and outperforms other baselines. Methods above NSG use more queries than NSG, while those below NSG use less (thus, NSG is an \emph{upper-bound}).} 
    \label{tab:main_table}
\end{table}

We also show that our method indeed learns disentangled representations regarding consistency. We conduct a similar experiment as in Sec.~\ref{sec:Naive} for CF-NSG, and show the result in the third row of Fig.~\ref{fig:entanglement}. Our CF-NSG relatively well preserves object-intrinsic properties regardless of the object's location. Making use of disentangled representations, we also can apply CF-NSG to various scene compositions in a more stable manner (see Supp. Material).

\noindent
\textbf{Ablation Study.}
In Tab.~\ref{tab:ablation}, we show the benefit of using score-based skipping and feature-reusing framework progressively. After skipping, 62.7\% of queries are left, while 75.8\% of them take the reusing pass. We also show our choice of 100 bins per dimension balances the trade-off between the rendered image quality and the memory cost. More ablations on various skipping and memory-efficient implementations are in Supp. Materials.

\begin{table}[t]
  \begin{minipage}[b]{0.45\textwidth}
  \centering
  \scalebox{0.8}{
  \begin{tabular}{l||c|c}
    \toprule
    {\parbox{1.5cm}{Method}} & {\parbox{1.3cm}{\centering $\#$Queries}} & {\parbox{1.3cm}{\centering PSNR($\uparrow$)}} \\ \midrule
    NSG\cite{ost2021neural} & 1 & 29.54 \\
    + Score-based skipping & 0.627 & 28.80 \\
    + \textbf{Feature-Reusing (CF-NSG)} & \textbf{0.152} & 28.70 \\ 
    \bottomrule
  \end{tabular}}
  \end{minipage}
  \hfill
  \begin{minipage}[b]{0.45\textwidth}
  \centering
  \scalebox{0.8}{
  \begin{tabular}{c||c|c}
    \toprule
    {\parbox{2.5cm}{\centering $\#$Bins per dim.}} & {\parbox{1.6cm}{\centering PSNR($\uparrow$)}} & {\parbox{1.6cm}{\centering Mem.(MB)}} \\
    \midrule
    75 & 28.59 & 165.73 \\
    100 & 28.70 & 313.83 \\
    125 & 28.72 & 503.12 \\
    \bottomrule
  \end{tabular}}
 \end{minipage}
  \vspace{3mm}
  \caption{\textbf{Ablations for each component and bin size.} We validate score-based skipping and feature-reusing frameworks respectively and evaluate effect of bin size.}  \label{tab:ablation}
\end{table}

\noindent
\textbf{Trade-off between Computation and Memory Cost.}
Since our approach stores and reuses consistent components, our efficiency gain is accompanied by additional memory footprint. Tab.~\ref{tab:trade-off} represents the rendering quality and additional memory cost as a function of computation cost in the number of queries and FLOPs. We observe a relatively small memory footprint of 300MB leads to large efficiency gains by up to 85\% less queries for forward pass, with little drop in the image quality. (The actual speed improvement in FLOPs is about 82.9\%, slightly less than 85\%, since reusing is still not completely free.)
In practice, a user may flexibly choose a proper reusing rate considering the resource budget.

\begin{table}[t]
  \centering
  \begin{adjustbox}{width=\textwidth}
  \scriptsize
  \begin{tabular}{c||c|c|c|c|c|c|c}
    \toprule
    {\parbox{1.2cm}{\centering \textbf{\#Queries}}} & 
    {\parbox{0.8cm}{\centering \textbf{0.15$\times$}}} & 
    {\parbox{0.8cm}{\centering \textbf{0.30$\times$}}} &
    {\parbox{0.8cm}{\centering \textbf{0.45$\times$}}} &
    {\parbox{0.8cm}{\centering \textbf{0.60$\times$}}} &
    {\parbox{0.8cm}{\centering \textbf{0.90$\times$}}} &
    {\parbox{0.8cm}{\centering \textbf{1$\times$}}} &
    {\parbox{1cm}{\centering \textbf{NSG(1$\times$)}}} \\ 
    FLOPs/frame & 8.78$\times10^{11}$ & 1.66$\times10^{12}$ & 2.30$\times10^{12}$ & 2.95$\times10^{12}$ & 4.50$\times10^{12}$ & 5.00$\times10^{12}$ & 5.12$\times10^{12}$ \\
    \midrule
    Mem.(MB) & 313.8 & 249.9 & 184.3 & 131.92 & 119.76 & 0 & 0 \\
    PSNR($\uparrow$) & 28.70 & 28.73 & 28.78 & 28.78 & 28.78 & 28.88 & 29.54 \\
    \bottomrule
  \end{tabular}
  \end{adjustbox}
  \vspace{.5mm}
  \caption{\textbf{Trade-off between speed and additional memory usage with CF-NSG.} We reveal relation between additional memory cost and number of queries for full feed-forward pass.}  \label{tab:trade-off}
\end{table}

\section{Related Work} \label{sec:Related}
Recently, the advancement of implicit or neural representations has enabled researchers to achieve photo-realistic views~\cite{jiang2020local,mescheder2019occupancy,niemeyer2020differentiable}. To suggest a better representation for over-smoothed renderings of the existing methods, Mildenhall \emph{et al.} \cite{mildenhall2020nerf} introduced Neural Radiance Fields (NeRF). However, training and rendering processes based on neural representations often require time-consuming ray marching. To improve efficiency of NeRF-based models, some research has introduced more efficient data structures \textit{e.g.,} caching~\cite{hedman2021baking,garbin2021fastnerf,mueller2022instant}, visual hull~\cite{kondo2021vaxnerf}, sparse voxel~\cite{liu2020neural}, view-dependenet multiplane image~\cite{wizadwongsa2021nex} and octree~\cite{yu2021plenoctrees}.

Another stream of research introduced representations of complex and dynamic scenes~\cite{li2021neural,pumarola2021d,park2021hypernerf}. To clarify the difference between static and dynamic scene representations, the latter involves movement in both the camera and the scene. Hence, it is important to consider interactions with the scene such as global illumination to determine the appearance of a dynamic object. Due to this difference, more factors need to be considered to efficiently render the dynamic scenes besides simply modeling each canonical space of dynamic object as an efficient radiance fields for static scenes \textit{i.e.,} directly applying ~\cite{mueller2022instant,wizadwongsa2021nex,kondo2021vaxnerf,hedman2021baking,yu2021plenoctrees,liu2020neural,garbin2021fastnerf}.

\section{Summary} \label{sec:Summary}
We propose CF-NSG, a novel framework for representing multi-object dynamic scenes efficiently by utilizing a feature-reusing framework based on consistency-fields. CF-NSG is able to render images using only 15\% of the original number of queries with little compromise to the image quality. Also CF-NSG enjoys more extended novel scene manipulation, taking advantage of disentangled representation with respect to spatio-temporal consistency.

\newpage

\section*{Acknowledgement}
This work was supported by the New Faculty
Startup Fund from Seoul National University and by National Research Foundation (NRF) grant (No. 2021H1D3A2A03038607/15\%, 2022R1C1C1010627/15\%) and Institute of Information \& communications Technology Planning \& Evaluation (IITP) grant (No. 2022-0-00264/10\%, 2021-0-01778/10\%, 2022-0-00320 /25\%, No.2022-0-00953/25\%) funded by the Korea government (MSIT).

\bibliography{main}

\begin{thebibliography}{28}
\providecommand{\natexlab}[1]{#1}
\providecommand{\url}[1]{\texttt{#1}}
\expandafter\ifx\csname urlstyle\endcsname\relax
  \providecommand{\doi}[1]{doi: #1}\else
  \providecommand{\doi}{doi: \begingroup \urlstyle{rm}\Url}\fi

\bibitem[Ahmadyan et~al.(2021)Ahmadyan, Zhang, Ablavatski, Wei, and
  Grundmann]{ahmadyan2021objectron}
Adel Ahmadyan, Liangkai Zhang, Artsiom Ablavatski, Jianing Wei, and Matthias
  Grundmann.
\newblock Objectron: A large scale dataset of object-centric videos in the wild
  with pose annotations.
\newblock In \emph{Proceedings of the IEEE/CVF Conference on Computer Vision
  and Pattern Recognition}, pages 7822--7831, 2021.

\bibitem[Chu et~al.(2020)Chu, Xie, Mayer, Leal-Taix{\'e}, and
  Thuerey]{chu2020learning}
Mengyu Chu, You Xie, Jonas Mayer, Laura Leal-Taix{\'e}, and Nils Thuerey.
\newblock Learning temporal coherence via self-supervision for {GAN}-based
  video generation.
\newblock \emph{ACM Transactions on Graphics (TOG)}, 39\penalty0 (4):\penalty0
  75--1, 2020.

\bibitem[Fan et~al.(2017)Fan, Su, and Guibas]{fan2017point}
Haoqiang Fan, Hao Su, and Leonidas~J Guibas.
\newblock A point set generation network for 3{D} object reconstruction from a
  single image.
\newblock In \emph{Proceedings of the IEEE conference on computer vision and
  pattern recognition}, pages 605--613, 2017.

\bibitem[Garbin et~al.(2021)Garbin, Kowalski, Johnson, Shotton, and
  Valentin]{garbin2021fastnerf}
Stephan~J Garbin, Marek Kowalski, Matthew Johnson, Jamie Shotton, and Julien
  Valentin.
\newblock Fast{N}e{RF}: High-fidelity neural rendering at 200fps.
\newblock In \emph{Proceedings of the IEEE/CVF International Conference on
  Computer Vision}, pages 14346--14355, 2021.

\bibitem[Geiger et~al.(2012)Geiger, Lenz, and Urtasun]{geiger2012we}
Andreas Geiger, Philip Lenz, and Raquel Urtasun.
\newblock Are we ready for autonomous driving? the {KITTI} vision benchmark
  suite.
\newblock In \emph{IEEE conference on computer vision and pattern recognition},
  pages 3354--3361. IEEE, 2012.

\bibitem[Hedman et~al.(2021)Hedman, Srinivasan, Mildenhall, Barron, and
  Debevec]{hedman2021baking}
Peter Hedman, Pratul~P Srinivasan, Ben Mildenhall, Jonathan~T Barron, and Paul
  Debevec.
\newblock Baking neural radiance fields for real-time view synthesis.
\newblock In \emph{Proceedings of the IEEE/CVF International Conference on
  Computer Vision}, pages 5875--5884, 2021.

\bibitem[Jiang et~al.(2020)Jiang, Sud, Makadia, Huang, Nie{\ss}ner, Funkhouser,
  et~al.]{jiang2020local}
Chiyu Jiang, Avneesh Sud, Ameesh Makadia, Jingwei Huang, Matthias Nie{\ss}ner,
  Thomas Funkhouser, et~al.
\newblock Local implicit grid representations for 3{D} scenes.
\newblock In \emph{Proceedings of the IEEE/CVF Conference on Computer Vision
  and Pattern Recognition}, pages 6001--6010, 2020.

\bibitem[Kondo et~al.(2021)Kondo, Ikeda, Tagliasacchi, Matsuo, Ochiai, and
  Gu]{kondo2021vaxnerf}
Naruya Kondo, Yuya Ikeda, Andrea Tagliasacchi, Yutaka Matsuo, Yoichi Ochiai,
  and Shixiang~Shane Gu.
\newblock Vax{N}e{RF}: Revisiting the classic for voxel-accelerated neural
  radiance field.
\newblock \emph{arXiv preprint arXiv:2111.13112}, 2021.

\bibitem[Li et~al.(2021)Li, Niklaus, Snavely, and Wang]{li2021neural}
Zhengqi Li, Simon Niklaus, Noah Snavely, and Oliver Wang.
\newblock Neural scene flow fields for space-time view synthesis of dynamic
  scenes.
\newblock In \emph{Proceedings of the IEEE/CVF Conference on Computer Vision
  and Pattern Recognition}, pages 6498--6508, 2021.

\bibitem[Liu et~al.(2020)Liu, Gu, Lin, Chua, and Theobalt]{liu2020neural}
Lingjie Liu, Jiatao Gu, Kyaw~Zaw Lin, Tat-Seng Chua, and Christian Theobalt.
\newblock Neural sparse voxel fields.
\newblock \emph{arXiv preprint arXiv:2007.11571}, 2020.

\bibitem[Mescheder et~al.(2019)Mescheder, Oechsle, Niemeyer, Nowozin, and
  Geiger]{mescheder2019occupancy}
Lars Mescheder, Michael Oechsle, Michael Niemeyer, Sebastian Nowozin, and
  Andreas Geiger.
\newblock Occupancy networks: Learning 3{D} reconstruction in function space.
\newblock In \emph{Proceedings of the IEEE/CVF Conference on Computer Vision
  and Pattern Recognition}, pages 4460--4470, 2019.

\bibitem[Mildenhall et~al.(2020)Mildenhall, Srinivasan, Tancik, Barron,
  Ramamoorthi, and Ng]{mildenhall2020nerf}
Ben Mildenhall, Pratul~P Srinivasan, Matthew Tancik, Jonathan~T Barron, Ravi
  Ramamoorthi, and Ren Ng.
\newblock Ne{RF}: Representing scenes as neural radiance fields for view
  synthesis.
\newblock In \emph{European conference on computer vision}, pages 405--421.
  Springer, 2020.

\bibitem[M\"uller et~al.(2022)M\"uller, Evans, Schied, and
  Keller]{mueller2022instant}
Thomas M\"uller, Alex Evans, Christoph Schied, and Alexander Keller.
\newblock Instant neural graphics primitives with a multiresolution hash
  encoding.
\newblock \emph{ACM Trans. Graph.}, 41\penalty0 (4):\penalty0 102:1--102:15,
  July 2022.

\bibitem[Niemeyer and Geiger(2021)]{niemeyer2021giraffe}
Michael Niemeyer and Andreas Geiger.
\newblock {GIRAFFE}: Representing scenes as compositional generative neural
  feature fields.
\newblock In \emph{Proceedings of the IEEE/CVF Conference on Computer Vision
  and Pattern Recognition}, pages 11453--11464, 2021.

\bibitem[Niemeyer et~al.(2020)Niemeyer, Mescheder, Oechsle, and
  Geiger]{niemeyer2020differentiable}
Michael Niemeyer, Lars Mescheder, Michael Oechsle, and Andreas Geiger.
\newblock Differentiable volumetric rendering: Learning implicit 3{D}
  representations without 3{D} supervision.
\newblock In \emph{Proceedings of the IEEE/CVF Conference on Computer Vision
  and Pattern Recognition}, pages 3504--3515, 2020.

\bibitem[Ost et~al.(2021)Ost, Mannan, Thuerey, Knodt, and Heide]{ost2021neural}
Julian Ost, Fahim Mannan, Nils Thuerey, Julian Knodt, and Felix Heide.
\newblock Neural scene graphs for dynamic scenes.
\newblock In \emph{Proceedings of the IEEE/CVF Conference on Computer Vision
  and Pattern Recognition}, pages 2856--2865, 2021.

\bibitem[Park et~al.(2019)Park, Florence, Straub, Newcombe, and
  Lovegrove]{park2019deepsdf}
Jeong~Joon Park, Peter Florence, Julian Straub, Richard Newcombe, and Steven
  Lovegrove.
\newblock Deep{SDF}: Learning continuous signed distance functions for shape
  representation.
\newblock In \emph{Proceedings of the IEEE/CVF Conference on Computer Vision
  and Pattern Recognition}, pages 165--174, 2019.

\bibitem[Park et~al.(2021)Park, Sinha, Hedman, Barron, Bouaziz, Goldman,
  Martin-Brualla, and Seitz]{park2021hypernerf}
Keunhong Park, Utkarsh Sinha, Peter Hedman, Jonathan~T Barron, Sofien Bouaziz,
  Dan~B Goldman, Ricardo Martin-Brualla, and Steven~M Seitz.
\newblock Hyper{N}e{RF}: A higher-dimensional representation for topologically
  varying neural radiance fields.
\newblock \emph{arXiv preprint arXiv:2106.13228}, 2021.

\bibitem[Pumarola et~al.(2021)Pumarola, Corona, Pons-Moll, and
  Moreno-Noguer]{pumarola2021d}
Albert Pumarola, Enric Corona, Gerard Pons-Moll, and Francesc Moreno-Noguer.
\newblock D-{N}e{RF}: Neural radiance fields for dynamic scenes.
\newblock In \emph{Proceedings of the IEEE/CVF Conference on Computer Vision
  and Pattern Recognition}, pages 10318--10327, 2021.

\bibitem[Sitzmann et~al.(2019)Sitzmann, Thies, Heide, Nie{\ss}ner, Wetzstein,
  and Zollhofer]{sitzmann2019deepvoxels}
Vincent Sitzmann, Justus Thies, Felix Heide, Matthias Nie{\ss}ner, Gordon
  Wetzstein, and Michael Zollhofer.
\newblock Deep{V}oxels: Learning persistent 3d feature embeddings.
\newblock In \emph{Proceedings of the IEEE/CVF Conference on Computer Vision
  and Pattern Recognition}, pages 2437--2446, 2019.

\bibitem[Skorokhodov et~al.(2021)Skorokhodov, Ignatyev, and
  Elhoseiny]{skorokhodov2021adversarial}
Ivan Skorokhodov, Savva Ignatyev, and Mohamed Elhoseiny.
\newblock Adversarial generation of continuous images.
\newblock In \emph{Proceedings of the IEEE/CVF Conference on Computer Vision
  and Pattern Recognition}, pages 10753--10764, 2021.

\bibitem[Suarez(2017)]{suarez2017language}
Joseph Suarez.
\newblock Language modeling with recurrent highway hypernetworks.
\newblock \emph{Advances in neural information processing systems}, 30, 2017.

\bibitem[Wang et~al.(2003)Wang, Simoncelli, and Bovik]{wang2003multiscale}
Zhou Wang, Eero~P Simoncelli, and Alan~C Bovik.
\newblock Multiscale structural similarity for image quality assessment.
\newblock In \emph{The Thrity-Seventh Asilomar Conference on Signals, Systems
  \& Computers, 2003}, volume~2, pages 1398--1402. IEEE, 2003.

\bibitem[Wizadwongsa et~al.(2021)Wizadwongsa, Phongthawee, Yenphraphai, and
  Suwajanakorn]{wizadwongsa2021nex}
Suttisak Wizadwongsa, Pakkapon Phongthawee, Jiraphon Yenphraphai, and Supasorn
  Suwajanakorn.
\newblock {N}e{X}: Real-time view synthesis with neural basis expansion.
\newblock In \emph{Proceedings of the IEEE/CVF Conference on Computer Vision
  and Pattern Recognition}, pages 8534--8543, 2021.

\bibitem[Yan et~al.(2016)Yan, Yang, Yumer, Guo, and Lee]{yan2016perspective}
Xinchen Yan, Jimei Yang, Ersin Yumer, Yijie Guo, and Honglak Lee.
\newblock Perspective transformer nets: Learning single-view 3{D} object
  reconstruction without 3{D} supervision.
\newblock \emph{arXiv preprint arXiv:1612.00814}, 2016.

\bibitem[Yu et~al.(2021)Yu, Li, Tancik, Li, Ng, and
  Kanazawa]{yu2021plenoctrees}
Alex Yu, Ruilong Li, Matthew Tancik, Hao Li, Ren Ng, and Angjoo Kanazawa.
\newblock Plenoctrees for real-time rendering of neural radiance fields.
\newblock In \emph{Proceedings of the IEEE/CVF International Conference on
  Computer Vision}, pages 5752--5761, 2021.

\bibitem[Zhang et~al.(2019)Zhang, Wang, and Zhang]{zhang2019vehicle}
Mingyang Zhang, Pengli Wang, and Xiaoman Zhang.
\newblock Vehicle color recognition using deep convolutional neural networks.
\newblock In \emph{Proceedings of the 2019 International Conference on
  Artificial Intelligence and Computer Science}, page 236–238. Association
  for Computing Machinery, 2019.

\bibitem[Zhang et~al.(2018)Zhang, Isola, Efros, Shechtman, and
  Wang]{zhang2018unreasonable}
Richard Zhang, Phillip Isola, Alexei~A Efros, Eli Shechtman, and Oliver Wang.
\newblock The unreasonable effectiveness of deep features as a perceptual
  metric.
\newblock In \emph{Proceedings of the IEEE conference on computer vision and
  pattern recognition}, pages 586--595, 2018.

\end{thebibliography}
\end{document}


\maketitle

\section{Illustration of the Static Model}

\begin{figure}[h!]
    \centering
    \includegraphics[width=1\linewidth]{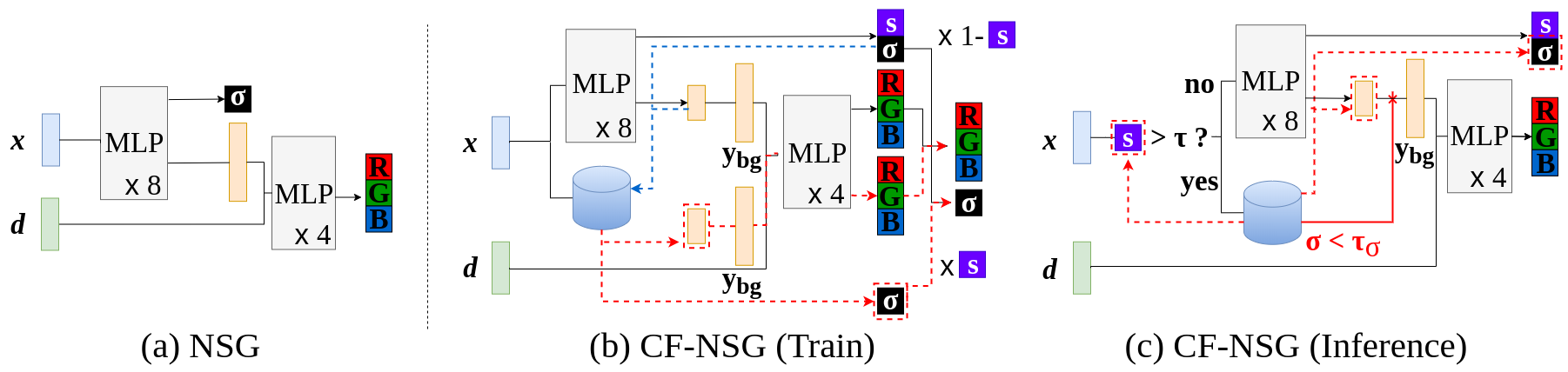}
    \caption{\textbf{The overall pipeline of CF-NSG (the static model).}}
    \label{fig:method_static}
\end{figure}

Fig.~\ref{fig:method_static} illustrates the overall pipeline for the static model of our CF-NSG. The static model receives 3D location $\textbf{\textit{x}}$ and viewing direction $\textbf{\textit{d}}$ in global coordinates as input and returns RGB and density values of the queries belonging to the static objects. The canonical features $\textbf{y}_{bg}$ are dependent on $\textbf{\textit{x}}$ and represent object-intrinsic properties. $\textbf{y}_{bg}$ and density $\sigma$ are stored in the memory bins during training and reused, while at inference queries are skipped, reused, or fully feed-forwarded based on the consistency scores $s_{bg}$ as same as the dynamic model.

\section{Pseudocode}
We provide a pseudocode for our CF-NSG in Alg.~\ref{algo:obj} (dynamic model) and in Alg.~\ref{algo:bg} (static model). Note that $o \in \mathcal{O}_c$ denotes each dynamic object belonging to a class $c \in \mathcal{C}$. Every query goes through either Alg.~\ref{algo:obj} or Alg.~\ref{algo:bg} based on whether it belongs to a dynamic object or the static background. Then, RGB and density values of the query are returned except when the query is likely to be in the empty-spaced bin and therefore skipped. After this step, the RGB and density values of all queries in each batch are integrated along the ray as in the original NeRF~\cite{mildenhall2020nerf}.

\begin{algorithm}[!p]
\DontPrintSemicolon
\KwInput{3D location $\textbf{\textit{x}}_{o}$, viewing direction $\textbf{\textit{d}}_{o}$, latent vector $\textbf{\textit{l}}_{o}$, global location $\textbf{\textit{p}}_{o}$ of dynamic object $o$ belonging to class $c$, 
the first 8-layers MLPs $\mathbf{F}_{c, 1}$, the second 4-layers MLPs $\mathbf{F}_{c, 2}$ of the dynamic model for class $c$,
bin memory $\mathcal{M}_{o}$ of the object $o$, hyperparameters $\tau$ and $\tau_{\sigma}$}

\textbf{[ Training ]} \\
\tcp{Going through the full forward pass}
$\textbf{y}_{o,\text{full}} \leftarrow \mathbf{F}_{c, 1}(\textbf{\textit{x}}_{o}, \textbf{\textit{l}}_{o})$ \\
$(r, g, b, \sigma, s_o)_{\text{full}} \leftarrow \mathbf{F}_{c, 2}(\textbf{y}_{o,\text{full}}, \textbf{\textit{d}}_o, \textbf{\textit{p}}_{o})$ \\
\textit{\small \textbf{Update}}($\mathcal{M}_{o}, \textbf{\textit{x}}_{o}, (\textbf{y}_o, \sigma, s_o)_{\text{full}})$ \\

\tcp{Going through the reusing pass}
\If {$\textit{\small \textbf{DoesMemoryExist}}(\mathcal{M}_{o}, \textbf{\textit{x}}_{o})$}
{$(\textbf{y}_o, \sigma, s_o)_{\text{reuse}} \leftarrow \textit{\small \textbf{Retrieve}}(\mathcal{M}_{o}, \textbf{\textit{x}}_{o})$ \\
$(r, g, b)_{\text{reuse}} \leftarrow \mathbf{F}_{c, 2}(\textbf{y}_{o, \text{reuse}}, \textbf{\textit{d}}_o, \textbf{\textit{p}}_{o})$ \\ 
$(r, g, b, \sigma)_{\text{interpolate}} \leftarrow  s_{o,\text{full}} \cdot (r, g, b, \sigma)_{\text{reuse}} + (1 - s_{o,\text{full}}) \cdot (r, g, b, \sigma)_{\text{full}}$}
\Else {return}
\KwOutput{$(r, g, b, \sigma)_{\text{interpolate}}$}

\textbf{[ Inference ]} \\
\If {$\textit{\small \textbf{DoesMemoryExist}}(\mathcal{M}_{o}, \textbf{\textit{x}}_{o})$}
{$(\textbf{y}_o, \sigma, s_o) \leftarrow \textit{\small \textbf{Retrieve}}(\mathcal{M}_{o}, \textbf{\textit{x}}_{o})$ \\
  \tcp{Skipping the whole pass}
  \If {$s_o > \tau \; and \; \sigma < \tau_{\sigma}$} {return}
  \tcp{Activate the reusing pass}
  \ElseIf {$s_o > \tau \; and \; \sigma \geq \tau_{\sigma}$}
  {$(r, g, b) \leftarrow \mathbf{F}_{c, 2}(\textbf{y}_o, \textbf{\textit{d}}_o, \textbf{\textit{p}}_{o})$}
  \tcp{Activate the full forward pass}
  \Else {$(\textbf{y}_o, \sigma, s_o) \leftarrow \mathbf{F}_{c, 1}(\textbf{\textit{x}}_{o}, \textbf{\textit{l}}_{o})$ \\
  $(r, g, b) \leftarrow \mathbf{F}_{c, 2}(\textbf{y}_o, \textbf{\textit{d}}_o, \textbf{\textit{p}}_{o})$}}

\Else {$(\textbf{y}_o, \sigma, s_o) \leftarrow \mathbf{F}_{c, 1}(\textbf{\textit{x}}_{o}, \textbf{\textit{l}}_{o})$ \\
$(r, g, b) \leftarrow \mathbf{F}_{c, 2}(\textbf{y}_o, \textbf{\textit{d}}_o, \textbf{\textit{p}}_{o})$}  
\KwOutput{$(r, g, b, \sigma)$}
\caption{\text{ Dynamic model of CF-NSG}} \label{algo:obj}
\end{algorithm}

\begin{algorithm}[p!]
\KwInput{3D location $\textbf{\textit{x}}$, viewing direction $\textbf{\textit{d}}$, the first 8-layers MLPs $\mathbf{F}_{bg, 1}$, the second 4-layers MLPs $\mathbf{F}_{bg, 2}$ of static model, bin memory $\mathcal{M}_{bg}$ for the static model, hyperparameters $\tau$ and $\tau_{\sigma}$}

\textbf{[ Training ]} \\
\tcp{Going through the full forward pass}
$\textbf{y}_{\text{full}} \leftarrow \mathbf{F}_{bg, 1}(\textbf{\textit{x}})$ \\
$(r, g, b, \sigma, s)_{\text{full}} \leftarrow \mathbf{F}_{bg, 2}(\textbf{y}_{\text{full}}, \textbf{\textit{d}})$ \\
\textit{\small \textbf{Update}}($\mathcal{M}_{bg}, \textbf{\textit{x}}, (\textbf{y}, \sigma, s)_{\text{full}})$ \\

\tcp{Going through the reusing pass}
\If {$\textit{\small \textbf{DoesMemoryExist}}(\mathcal{M}_{bg}, \textbf{\textit{x}})$}
{$(\textbf{y}, \sigma, s)_{\text{reuse}} \leftarrow \textit{\small \textbf{Retrieve}}(\mathcal{M}_{bg}, \textbf{\textit{x}})$ \\
$(r, g, b)_{\text{reuse}} \leftarrow \mathbf{F}_{bg, 2}(\textbf{y}_{\text{reuse}}, \textbf{\textit{d}})$ \\ 
$(r, g, b, \sigma)_{\text{interpolate}} \leftarrow  s_{\text{full}} \cdot (r, g, b, \sigma)_{\text{reuse}} + (1 - s_{\text{full}}) \cdot (r, g, b, \sigma)_{\text{full}}$}
\Else {return}
\KwOutput{$(r, g, b, \sigma)_{\text{interpolate}}$}

\textbf{[ Inference ]} \\
\If {$\textit{\small \textbf{DoesMemoryExist}}(\mathcal{M}_{bg}, \textbf{\textit{x}})$}
{$(\textbf{y}, \sigma, s) \leftarrow \textit{\small \textbf{Retrieve}}(\mathcal{M}_{bg}, \textbf{\textit{x}})$ \\
  \tcp{Skipping the whole pass}
  \If {$s > \tau \; and \; \sigma < \tau_{\sigma}$} {return}
  \tcp{Activate the reusing pass}
  \ElseIf {$s > \tau \; and \; \sigma \geq \tau_{\sigma}$}
  {$(r, g, b) \leftarrow \mathbf{F}_{bg, 2}(\textbf{y}, \textbf{\textit{d}})$}
  \tcp{Activate the full forward pass}
  \Else {$(\textbf{y}, \sigma, s) \leftarrow \mathbf{F}_{bg, 1}(\textbf{\textit{x}})$ \\
  $(r, g, b) \leftarrow \mathbf{F}_{bg, 2}(\textbf{y}, \textbf{\textit{d}})$}}

\Else {$(\textbf{y}, \sigma, s) \leftarrow \mathbf{F}_{bg, 1}(\textbf{\textit{x}})$ \\
$(r, g, b) \leftarrow \mathbf{F}_{bg, 2}(\textbf{y}, \textbf{\textit{d}})$}  
\KwOutput{$(r, g, b, \sigma)$}
\caption{\text{ Static model of CF-NSG}} \label{algo:bg}
\end{algorithm}

\section{Implementation Details}
As we want our model to learn consistency based on the understanding of color and volumetric density of the scene, we warm up CF-NSG using only NSG loss terms (the first and last term in Eq.~\eqref{eq:Ours}) and afterwards train CF-NSG with Eq.~\eqref{eq:Ours}. For KITTI~\cite{geiger2012we}, we pretrain CF-NSG for 275k iterations and finetune for 75k iterations, while for Objectron~\cite{ahmadyan2021objectron} we use 500k and 235k iterations, respectively. We use a single Nvidia TITIAN Xp GPU for training and find the setting with the number of bins for each coordinate axis $N = 100$, score threshold $\tau = 0.5$, density threshold $\tau_{\sigma} = 0.9$, $l = 256$ ($m = 4$) and coefficient of score regularization term $\lambda = 10^{-8}$ work well for our experiments.

Empirically, we find that for the static model for KITTI, storing color directly to the bins instead of canonical features does not harm the image quality. Thanks to the learnable score, changeable parts of the background are well captured and return low scores, therefore rendered from the full forward pass. Instead of storing $l$-dimensional canonical features, using 3-dimensional RGB enhances the efficiency in terms of memory cost.

Backgrounds of the Objectron dataset are mostly composed of floors and walls that are spatially close to the object; that is, the backgrounds tend to include both vertical and horizontal planes of the object. Therefore, modeling the background with parallel 2D planes~\cite{zhou2018stereo}, following NSG, leads to an inappropriate fitting. As shown in Fig.~\ref{fig:various_objectron}, NSG results in a blurred background. When we finetune our CF-NSG after pretraining NSG loss, the static model cannot learn consistency scores and canonical features stably because incorrect RGB color estimation produces rendered images that cannot give an accurate instruction about consistency. Thus, we do not use a feature-reusing framework for the static model and concentrate on the dynamic model for a comparison on Objectron. We also report the number of queries that belong to the object in Tab.~\ref{tab:main_table} in the main manuscript.

Across all the experiments, we use the official implementation of NSG~\cite{ost2021neural} for NSG and our CF-NSG: we define 6 planes for the static model and calculate the intersections of each plane and each ray, instead of performing ray marching. For the dynamic model, we use 7 sampling points on each ray which intersects with the bounding box of each of the dynamic objects. Following the above settings, we sample 13 points in total on each ray and do not use an additional fine network for NeRF+time~\cite{ost2021neural} in Tab.~\ref{tab:main_table} (in the main manuscript). For NeRF~\cite{mildenhall2020nerf} and NSVF~\cite{liu2020neural}, on the other hand, we use their official implementation and utilize both coarse and fine network using 64 and 128 sampling points per ray, respectively. For `NSG-reduced' on both KITTI and Objectron, we reduce the number of queries until objects in a rendered image are severely damaged and indistinguishable from the background. 

The optimization of NSG, NeRF, and NeRF+time takes 350k iterations for KITTI and 735k iterations for Objectron, respectively, while that of D-NeRF~\cite{pumarola2021d} and NSVF takes 800k and 150k iterations for KITTI, respectively. We set initial voxel size as 0.5 for NSVF to fit in our budget.

\section{Additional Results of CF-NSG}
\begin{figure}[h!]
  \centering
  \includegraphics[width=1\linewidth]{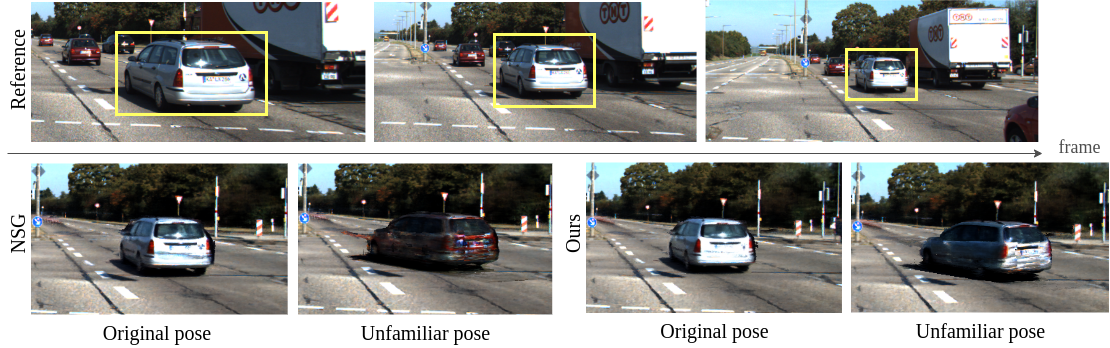}
  \caption{\textbf{Results on novel pose synthesis (rotation).}}
  \label{fig:dance}
\end{figure}

Making use of disentanglement between object-intrinsic and environmental representation, we apply CF-NSG to various scene compositions more stably. In Fig.~\ref{fig:dance}, we apply rotations on a dynamic object over a range of unfamiliar angles at which the appearance of the object is not entirely new but appear rarely in the training set. NSG properly represents the object for the frequently appeared views at training. For views beyond a certain range, however, NSG produces significantly irrelevant colors (see the left two images in the lower row of Fig.~\ref{fig:dance}), probably due to its entangled representations. On the other hand, CF-NSG fully understands the consistent properties (\textit{i.e.}, vehicle color), and therefore represents objects with significantly less deterioration.

\begin{figure}[h!]
    \centering
    \includegraphics[width=1\linewidth]{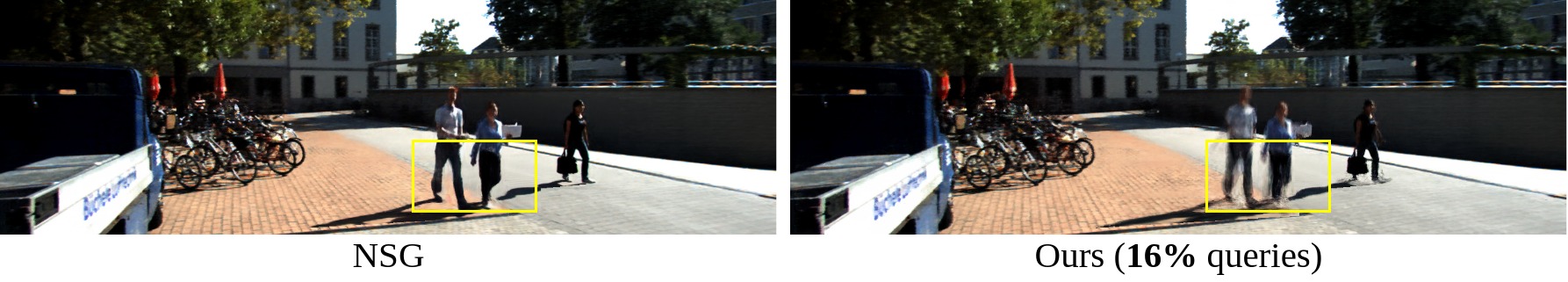}
    \caption{\textbf{A failure case of CF-NSG with elusiveness.}}
    \label{fig:failure}
\end{figure}

However, we show that there is still room for improvement in representing scenes, including fast movement and deformable objects. Since CF-NSG considers consistent properties of the objects in the scene, too rapid change in shape or location may present elusiveness. For instance, Fig.~\ref{fig:failure} represents CF-NSG's confusion when a pedestrian's pose changes largely. As a future work, it will be interesting to see if some prior knowledge about the object or finer adjustments to the consistency score helps to expand the spectrum of applicable scenes.

\begin{figure}[hbtp]
     \centering
     \includegraphics[width=1\linewidth]{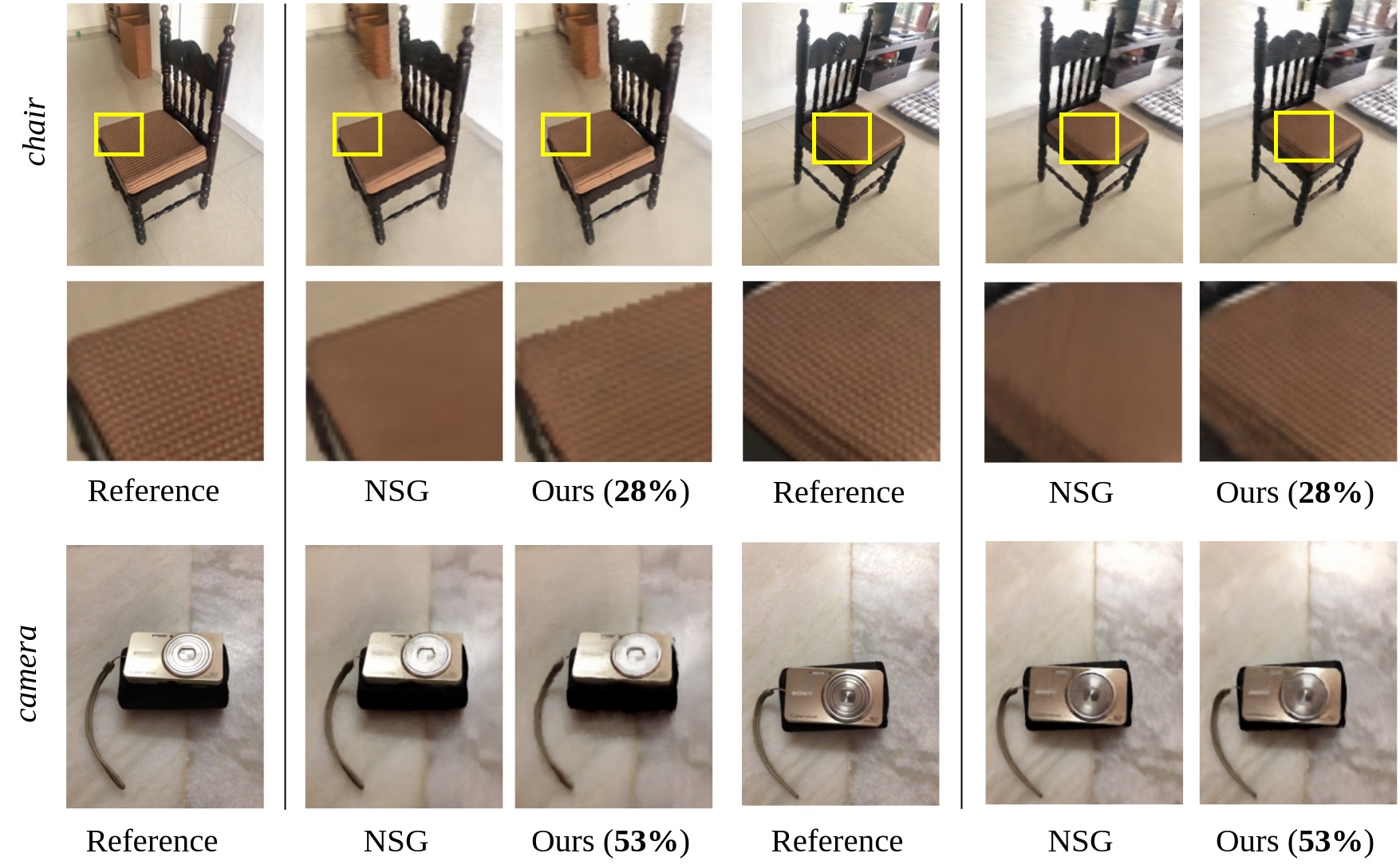}
     \caption{\textbf{Qualitative results on Objectron~\cite{ahmadyan2021objectron} reconstructing the reference frame.} CF-NSG uses only 28--53\% queries NSG uses to render an image.}
     \label{fig:various_objectron}
\end{figure}

\begin{figure}[hbtp]
     \centering
     \includegraphics[width=1\linewidth]{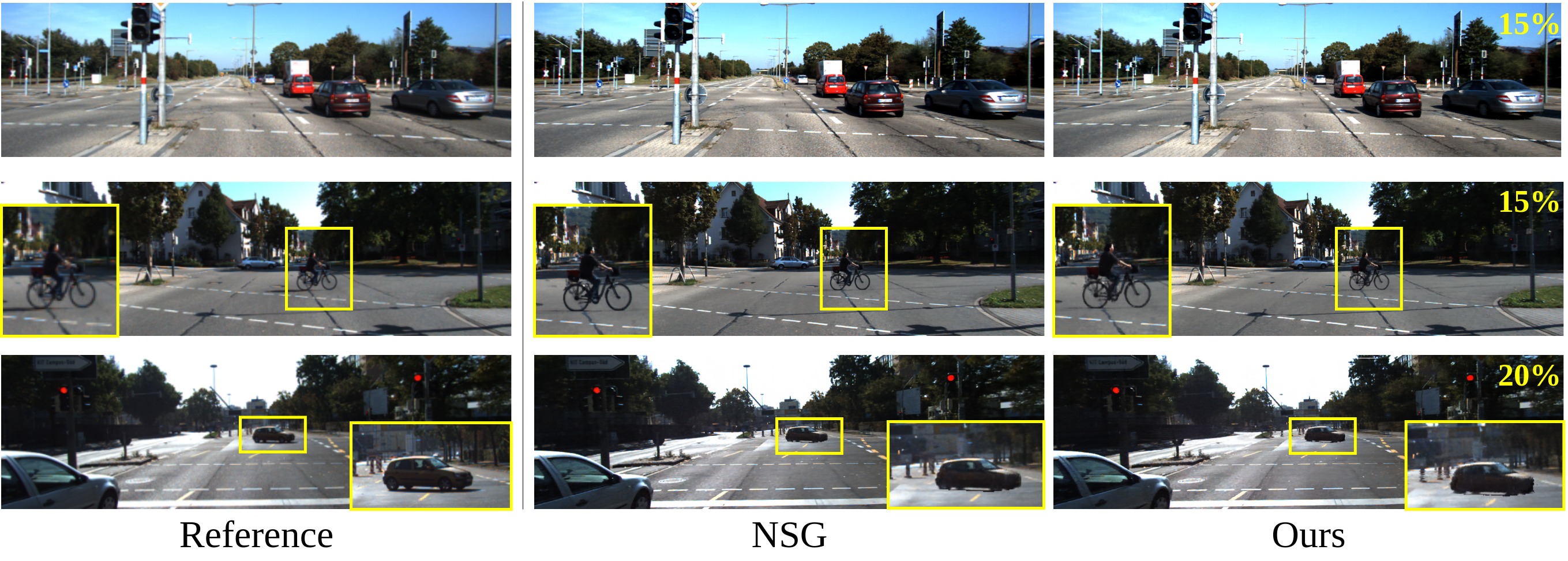}
     \caption{\textbf{Qualitative results on KITTI~\cite{geiger2012we} reconstructing the reference frame.} CF-NSG uses only 15--20\% queries NSG uses to render an image.}
     \label{fig:various_kitti}
\end{figure}

\begin{figure}[hbtp]
     \centering
     \includegraphics[width=1\linewidth]{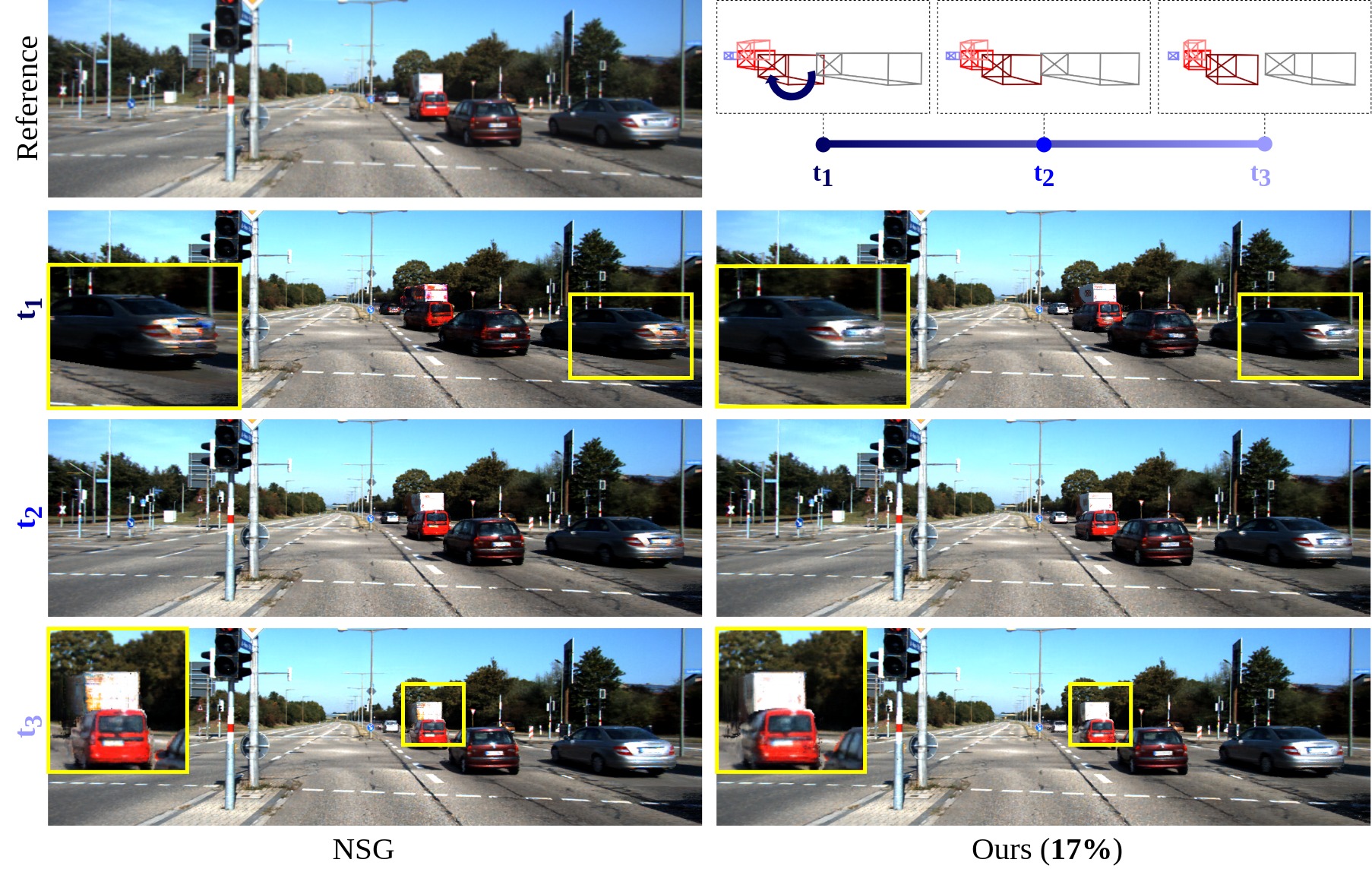}
     \caption{\textbf{Qualitative results on KITTI~\cite{geiger2012we} translating the objects.}}
     \label{fig:various_novel}
\end{figure}

\begin{figure}[hbtp]
     \centering
     \includegraphics[width=1\linewidth]{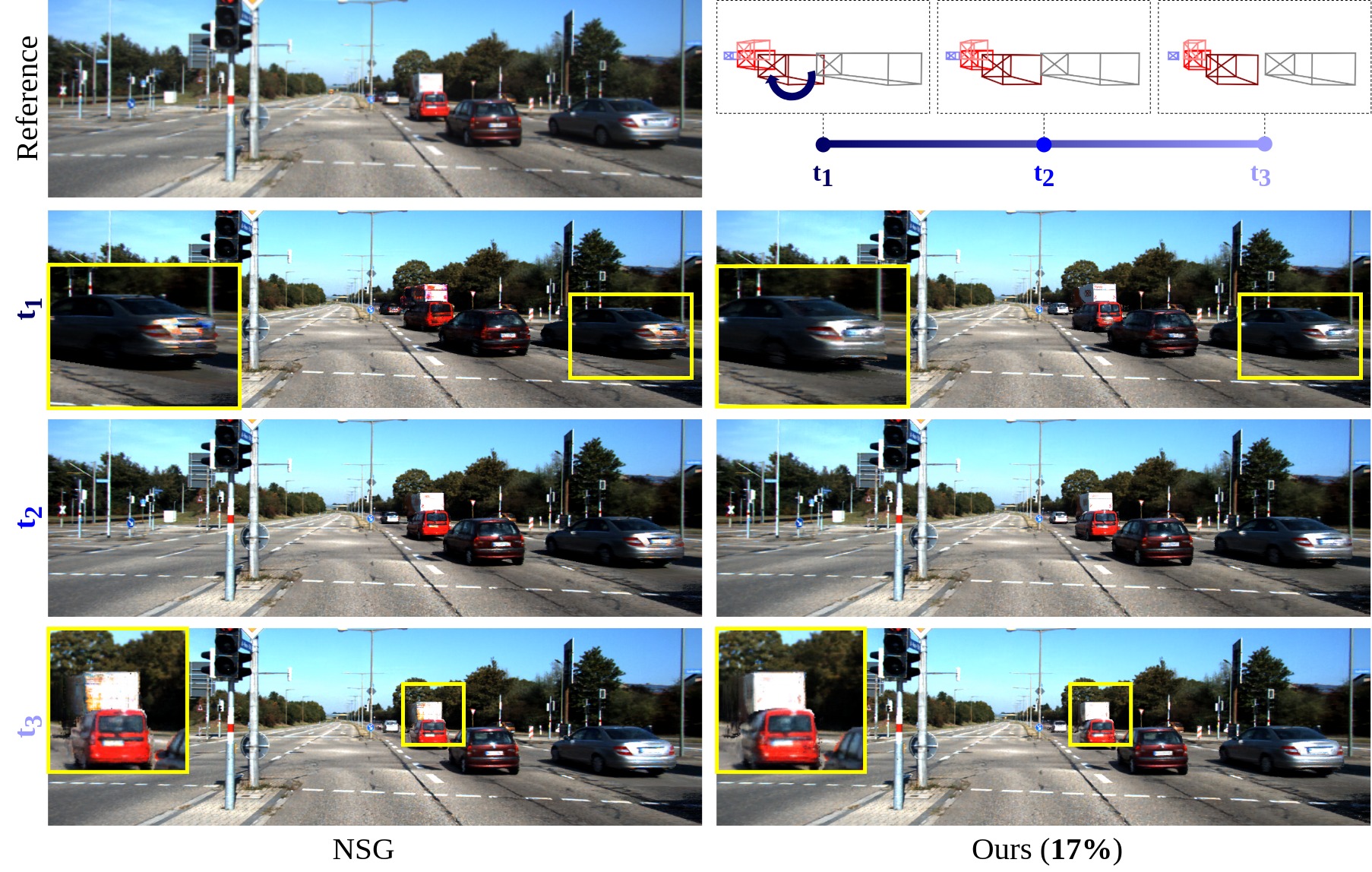}
     \caption{\textbf{Qualitative results on KITTI~\cite{geiger2012we} rotating the objects.}}
     \label{fig:various_novel2}
\end{figure}

Nevertheless, our CF-NSG can be applied to various scenes in KITTI~\cite{geiger2012we} and Objectron~\cite{ahmadyan2021objectron} including diverse objects and drastic camera view changes. We provide additional qualitative results for our CF-NSG in Fig.~\ref{fig:various_objectron}--\ref{fig:various_novel2}. Fig.~\ref{fig:various_objectron} and~\ref{fig:various_kitti} show the results of reconstructing the reference frame seen during the training. CF-NSG can be applied to various scenes with diverse objects, showing minimal degradation of image quality or in some cases (\textit{e.g.,} chair in Fig.~\ref{fig:various_objectron}) even being better at capturing the fine details of the object than NSG such as patterns on the seat cushion of the chair in spite of using only 15--53\% of queries. In Fig.~\ref{fig:various_novel} and~\ref{fig:various_novel2}, using dynamic objects in the reference frame, we manipulate a scene in a novel arrangement, \textit{e.g.,} translating the objects horizontally or rotating the objects on their global position. Thanks to disentanglement between object-intrinsic and environmental representations, CF-NSG performs an extended degree of scene manipulation more stably without abnormalities in vehicle color or shape.

\section{Additional Ablations}
In this section, we conduct additional ablation studies to verify: 1) our score-based skipping compared to density-only-based skipping, and 2) our memory-efficient implementation to see if it is effective in terms of both memory cost and preservation of image quality.

\begin{figure}[h]
  \centering
  \includegraphics[width=1\linewidth]{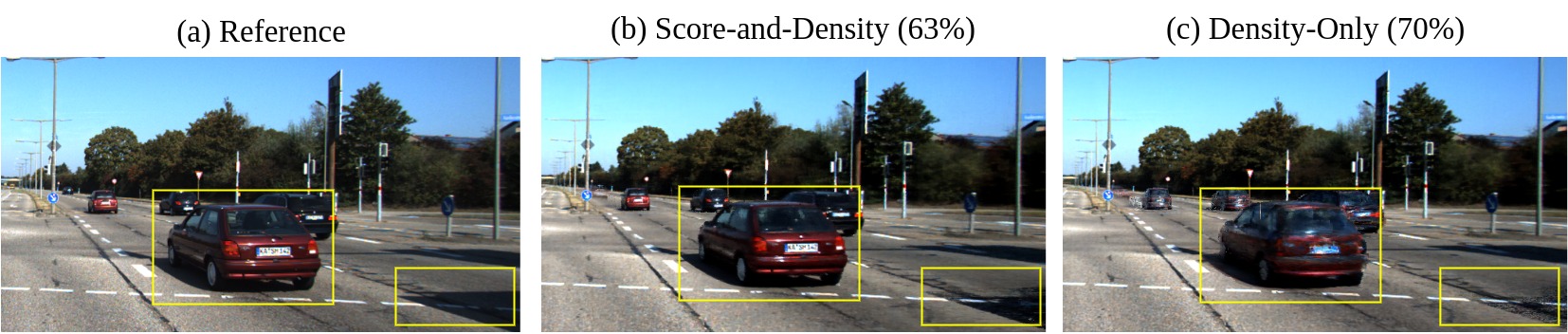}\\
  \caption{\textbf{Ablations for Skipping \textit{with} and \textit{without} Score.}}
  \label{fig:ablation_skipping}
\end{figure}

\vspace{0.1cm} \noindent
\textbf{Score-based Skipping.}
In Fig.~\ref{fig:ablation_skipping}, we reconstruct the reference scene in the training set using both the consistency score and density (b), or using the density only (c). Without the score, inappropriate skipping leaves black marks where the car windows should appear (yellow rectangle in the center in Fig.~\ref{fig:ablation_skipping}(c)). Since almost transparent windows have density values close to zero, using density alone leads to indistinctiveness between windows and the empty space. Therefore, most of the car windows are skipped erroneously. Also, the queries belonging to the shadow on the road are improperly skipped (yellow rectangle on the right corner in Fig.~\ref{fig:ablation_skipping}(c)). On the other hand, considering the consistency scores along with the density results in appropriate skipping with a minimal degradation in image quality, as shown in Fig.~\ref{fig:ablation_skipping}(b).

\vspace{0.1cm} \noindent
\textbf{Memory-efficient Implementation.}
In Tab.~\ref{tab:ablation_memory}, we report the memory usage and rendering quality metrics with the following three approaches: 1) implementation without any memory-efficient methods, 2) encoder-decoder architecture, and 3) low-rank factorization (used in our CF-NSG). For the encoder-decoder architecture, we feed $\mathbf{y} \in \mathbb{R}^{l}$ to the encoder and store the output $\mathbf{y}_{\text{intermediate}} \in \mathbb{R}^{n}$, where $n = l/4$. Then, we feed $\mathbf{y}_{\text{intermediate}}$ to the decoder and reuse the output $\mathbf{y}_{\text{out}} \in \mathbb{R}^{l}$ from the decoder. We use 1-layer MLP for both encoder and decoder, while adding ReLU activation only for the encoder. Tab.~\ref{tab:ablation_memory} shows that low-rank factorization is beneficial to both efficiency and image quality.

\begin{table}[h!]
    \centering
	\resizebox{1\textwidth}{!}{
	\begin{tabular}{c|c||c|c|c|c|c}
	\toprule
	{\parbox{1.5cm}{\centering \small Method}} & {\parbox{1.5cm}{\centering \small Mem.(MB)}} & {\parbox{1.6cm}{\centering \small PSNR($\uparrow$)}} & {\parbox{1.6cm}{\centering \small SSIM~\cite{wang2003multiscale}($\uparrow$)}} & {\parbox{1.6cm}{\centering \small LPIPS~\cite{zhang2018unreasonable}($\downarrow$)}} & {\parbox{1.6cm}{\centering \scriptsize tOF~\cite{chu2020learning}$\times10^6$($\downarrow$)}} & {\parbox{1.6cm}{\centering  \scriptsize tLP~\cite{chu2020learning}$\times100$($\downarrow$)}} \\
    \midrule 
    Without any & 5665.51 & 27.73 & 0.860 & \textbf{0.128} & 0.936 & 0.445
    \\
    Enc-Dec & 503.12 & 19.00 & 0.782 & 0.195 & 2.789 & 7.016 \\
    Low-rank & \textbf{313.83} & \textbf{28.70} & \textbf{0.891} & 0.204 & \textbf{0.766} & \textbf{0.266} \\
	\bottomrule
	\end{tabular}}
\vspace{.1mm}
\caption{\centering \textbf{Ablation for memory-efficient implementation.}} 
\label{tab:ablation_memory}
\end{table}

\clearpage
%
%

\bibliography{supplement}